%% file: main.tex
\documentclass[a4paper]{article}

\usepackage[english]{babel}
\usepackage[T1]{fontenc}

\usepackage[a4paper,top=3cm,bottom=2cm,left=3cm,right=3cm,marginparwidth=1.75cm]{geometry}

\usepackage[utf8]{inputenc} % allow utf-8 input
\usepackage[T1]{fontenc}    % use 8-bit T1 fonts
\usepackage{url}            % simple URL typesetting
\usepackage{booktabs}       % professional-quality tables
\usepackage{amsfonts}       % blackboard math symbols
\usepackage{nicefrac}       % compact symbols for 1/2, etc.
\usepackage{microtype}      % microtypography
\usepackage{textcomp}
\usepackage{siunitx}
\usepackage{comment}
\usepackage{multirow}
\usepackage{mathtools}
\usepackage{booktabs} % For formal tables
\usepackage{filecontents} %<-- To create the data file on the go
 \usepackage[ruled,vlined]{algorithm2e}
\usepackage{textcomp}
\usepackage{siunitx}
\usepackage{graphicx}  %Required
\usepackage{comment}
\usepackage{multirow}
\usepackage{booktabs} % For formal tables
\usepackage{filecontents} %<-- To create the data file on the go
 \usepackage[ruled,vlined]{algorithm2e}
\usepackage{amsmath}
\usepackage{amsfonts}
\usepackage{array}
\usepackage{bm}
\usepackage{caption,subcaption}
\usepackage{pgfplots}
\usepackage{pgfplotstable}

\newcommand{\T}[1]{\ensuremath{\mathcal{#1}}} % tensor
\newcommand{\M}[1]{\ensuremath{\bm{#1}}} % matrix
\newcommand{\V}[1]{\ensuremath{\bm{#1}}} % vector
\newcommand{\overbar}[1]{\mkern 1.5mu\overline{\mkern-1.5mu#1\mkern-1.5mu}\mkern 1.5mu}

\usepackage{url}
\usepackage{xspace,csquotes}
\usepackage{enumitem}

\newcommand{\mname}{\texttt{TASTE}\xspace}
\newcommand{\kp}[1]{{\color{blue} [KP: #1]}}
\newcommand{\ho}[1]{{\color{orange} [H: #1]}}
\newcommand{\js}[1]{\textcolor{magenta}{\emph{[JS: #1]}}}
\newcommand{\ari}[1]{{\color{red} [Ari: #1]}}

\newcommand{\hide}[1]{}

\newcommand\norm[1]{\left\lVert#1\right\rVert}
\begin{document}
\title{\mname: Temporal and Static Tensor Factorization for Phenotyping Electronic Health Records}

\author{ Ardavan Afshar$^1$, Ioakeim Perros$^2$, Haesun Park$^1$ \\ Christopher deFilippi$^3$, Xiaowei Yan$^4$, Walter Stewart, Joyce Ho$^5$, Jimeng Sun$^1$\\
$^1$Georgia Institute of Technology, ,$^2$Health at Scale $^3$INOVA,\\
$^4$Sutter Health, $^5$Emory University \\
}

\maketitle

\begin{abstract}
{\it Phenotyping electronic health records (EHR)} focuses on defining meaningful patient groups (e.g., heart failure group and diabetes group)  and identifying the temporal evolution of patients in those groups. Tensor factorization has been an effective tool for phenotyping. Most of the existing works assume either a static patient representation with aggregate data or only model temporal data. However, real EHR data contain both temporal (e.g., longitudinal clinical visits) and static information (e.g., patient demographics), which are difficult to model simultaneously. In this paper, we propose {\it T}emporal {\it A}nd {\it S}tatic {\it TE}nsor factorization (\mname) that jointly models both static and temporal information to extract phenotypes. %and their temporal evolution. \kp{the previous sentence only talks about the temporal evolution, but leaves unclear how static phenotypes are extracted}
\mname combines the PARAFAC2 model with non-negative matrix factorization to model a temporal and a static tensor. %\kp{i think the methodological contribution is vague; could we make this more specific?}
To fit the proposed model, we transform the original problem into simpler ones which are optimally solved in an alternating fashion. For each of the sub-problems, our proposed mathematical re-formulations lead to efficient sub-problem solvers. %\kp{I would change the previous sentence to: To fit the proposed model, we transform the original problem into smaller ones which are optimally solved in an alternating fashion. For each of the sub-problems, our proposed mathematical re-formulations lead to efficient sub-problem solvers.} \kp{Ari, please make sure that the "optimal" is correct.}
Comprehensive experiments on large EHR data from a heart failure (HF) study confirmed that \mname is up to $14 \times$ faster than several baselines and the resulting phenotypes were confirmed to be clinically meaningful by a cardiologist. Using 80 phenotypes extracted by \mname, a simple logistic regression can achieve the same level of area under the curve (AUC) for HF prediction compared to a deep learning model using recurrent neural networks (RNN) with 345 features. %\kp{Should we emphasize what the benefit is? e.g., comparable predictive power, but much easier to interpret?}
\end{abstract}
%\keywords{Tensor Factorization, Unsupervised Learning, Computational Phenotyping}

\input{01-introduction}

\input{02-related_work}
\input{03-method}

\input{04-experiments}

\input{05-Conclusion}
\input{06-Acknowledgements}
\bibliographystyle{unsrt}
\bibliography{main.bib} 

\input{06-suplemntary_matrials}

\end{document}

%% file: 01-introduction.tex
\section{\textbf{Introduction}}
Phenotyping is about identifying patient groups sharing similar clinically-meaningful characteristics and is essential for treatment development and management~\cite{richesson2016clinical}. However, the complexity and heterogeneity of the underlying patient information render a manual phenotyping impractical for large populations or complex conditions. Unsupervised EHR-based phenotyping based on tensor factorization, e.g., ~\cite{ho2014marble,Perros2018kdd,YQCFP19aaai}, provides effective alternatives. However, existing unsupervised phenotyping methods are unable to handle both static and dynamically-evolving information, which is the focus of this work.\\ 

\begin{figure}[htb]
\centering
\includegraphics[scale=0.35]{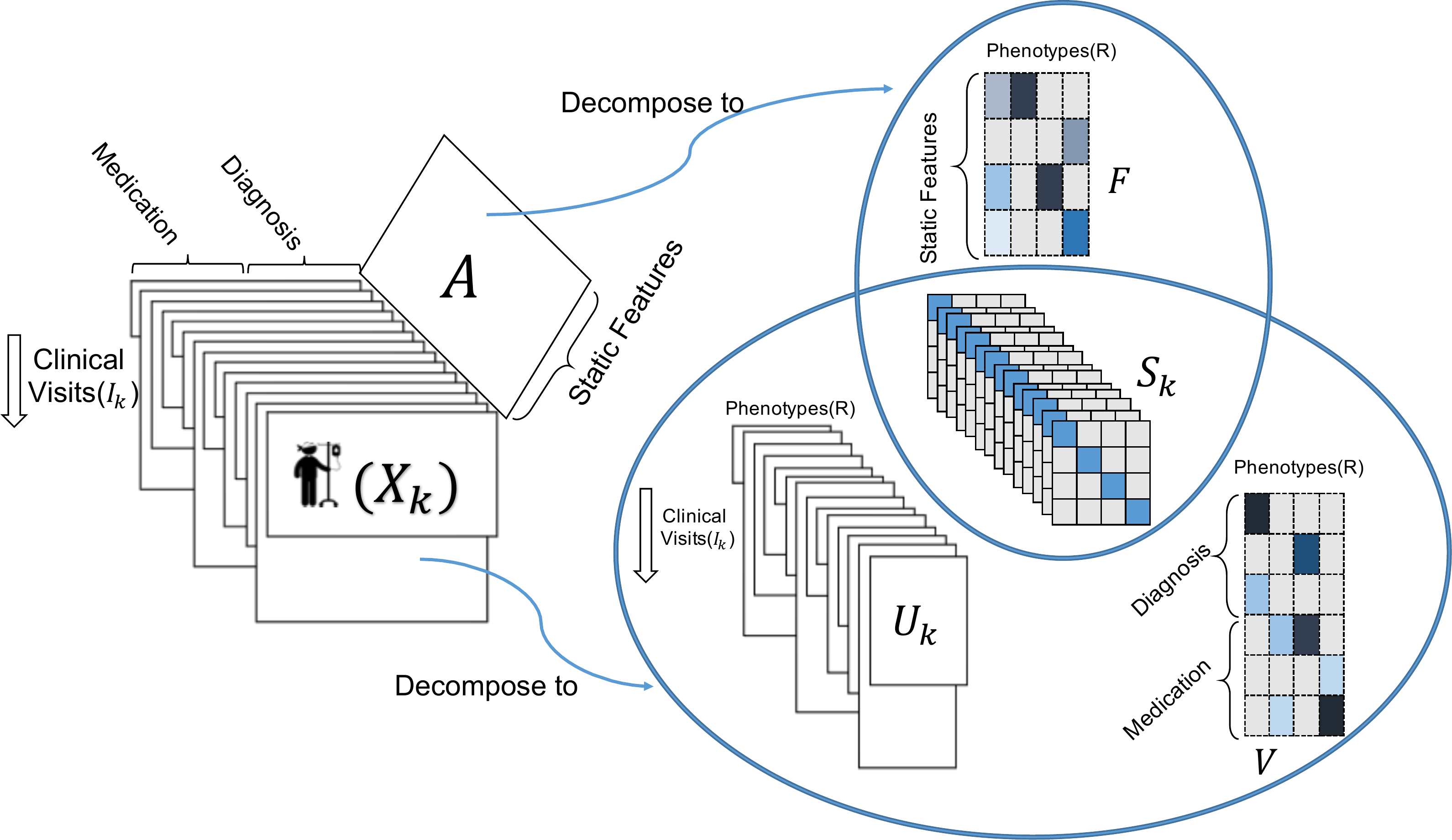}
\caption{ 
\mname applied on dynamically-evolving structured EHR data and static patient information. Each \M{X_k} represents the medical features recorded for different clinical visits for patient k.  Matrix \M{A} includes the static information (e.g., race, gender) of patients. \mname decomposes $\{ \M{X_k} \}$ into three parts: $\{\M{U_k}\}$, $\{\M{S_k}\}$, and \M{V}. Static matrix \M{A} is decomposed into two parts: $\{\M{S_k}\}$ and \M{F}. Note that $\{\M{S_k}\}$ (personalized phenotype scores) is shared between static and dynamically-evolving features. %\js{Keep in mind to make figures large enough to be legible  }
}
\label{fig_Coupled_PARAFAC2}
\end{figure}
Traditional tensor factorization models \cite{carroll1970analysis,hitchcock1927expression,harshman1970foundations}  assume the same dimensionality along with each tensor mode.  However, in practice one mode such as time can be irregular, e.g., different patients may vary by the number of clinical visits over time. 
To handle such longitudinal datasets, \cite{Perros2017-dh} and  \cite{afshar2018copa}  propose algorithms fitting the PARAFAC2 model~\cite{Hars1972b} which are faster and more scalable for handling irregular and sparse data. 
%Also, \cite{afshar2018copa} incorporates different constraints  such as temporal smoothness, sparsity to improve the interpretability. Still, 
However, these PARAFAC2 approaches only focus on modeling the dynamically-evolving features for every patient (e.g., the structured codes recorded for every visit). \textit{Static features} (such as race and gender) which do not evolve are completely neglected; yet, they are among  important information in phenotyping analyses (e.g., some diseases have the higher prevalence in a certain race).
%\kp{Do we have a result from our case study where Chris said: "that is a phenotype where the static feature info really makes sense?" can we use that here as an example?}
%Enabling to model both the temporally-evolving and static patient information is fundamental for \textit{phenotyping} based on EHRs, our main motivating application. \kp{i think summarizing what is the expected output here would be effective (and also connecting with Figure 1 as we do):  Overall, we are interested in extracting: a) clinically-meaningful groups of patients, b) justification of those groups based on both dynamically-evolving and static features; c) temporal trends indicating the trajectories of dynamically-evolving feature groups over time.} 
%To tackle the challenges above, we propose \mname that models both dynamic and static features for unsupervised knowledge extraction. 

To address this problem,  we propose a scalable method called \mname which is able to  jointly model both temporal and static features by combining the PARAFAC2 model with non-negative matrix factorization as shown in Figure~\ref{fig_Coupled_PARAFAC2}. We reformulate the original non-convex problem into simpler sub-problems (i.e., orthogonal Procrustes, least square and non-negativity constrained least square) and solve each of them efficiently by  avoiding  unnecessary computations (e.g., expensive Khatri-Rao products).  
%\kp{I am afraid the reader will think that there is no methodological novelty if we mention absolutely nothing about the method in both Abstract and Intro. Can't we make this more clear? at the very least, i would expect to read a couple of sentences in the first bullet point below, related to the methodological contribution (i.e., what is going on underneath, not merely the fact that it is a new technique)}

 %Overall, we are interested in extracting:  a) \textbf{patient groups}: clinically-meaningful groups of patients, b) \textbf{phenotype definitions}: dynamically-evolving and static features capturing similar characteristics for each group of patients and c) \textbf{phenotype trends}: temporal trends indicating  the trajectories of phenotypes over hospital visits.
We summarize our contributions below:
\begin{itemize}
\item \textbf{Temporal and Static Tensor Factorization:} % \kp{We propose a novel objective, which enables handling static features and preserves non-negativity of model factors, as well as constraints promoting model uniqueness.}
We formulate a new technique which jointly models static and dynamic features from EHR data as nonnegative factor matrices. 
%We formulate  a new coupled non-negative PARAFAC2 model, which enables us to jointly model static and dynamic features from EHR data as non-negative factor matrices. % modeling static and dynamic features from EHR data where all factor matrices  can be non-negative.
 %We formulate a new technique to jointly model static and dynamic features from EHR data as nonnegative factor matrices. 
\item \textbf{Fast and Accurate Algorithm:}
Our proposed fitting algorithm is up to $14~\times$ faster than the state of the art baseline. At the same time, \mname preserves model constraints which promote model uniqueness better than baselines while maintaining interpretability. %\js{it is not clear what model constraints are, and doesn't baseline provide that as well?   }%\kp{maybe this last sentence can be omitted since now we refer to uniqueness in the model contribution above. Also, we should also refer here to the projection, as i did in the abstract.} %\kp{this seems vague here, maybe directly refer to uniqueness, e.g.,: constraints which promote model uniqueness, thus boosting interpretability.}
%\item \textbf{Predicting the early detection of heart failure:} \mname is able to predict that whether a new patient will be diagnosed by heart failure or not. We show that by only 80 phenotypes, we can achieve a same prediction score (AUC) as a baseline with 935 features.  %\kp{let's decide if we want to include this study overall. my concern is that we need to have a good way of preventing arguments of the following type: "But, did you individually evaluate the interpretability of those 60 phenotypes? If you didn't, how do we know that this is more interpretable than 935 simpler features?}

%\kp{i doubt that this should be listed as one of the main contributions; it is not the main purpose of the paper, the results are not impressive and overall the way we present them in Table 3 are mostly to conclude that static information boosts predictive capability. i am afraid that if we emphasize it that much in the intro, people may try to kill the paper based on that.}

\item \textbf{Case Study on Heart Failure Phenotyping: }
%Heart failure (HF) phenotyping is an unsolved clinical problem. 
We demonstrate the practical impact of \mname through a case study on heart failure (HF) phenotyping. We identified clinically-meaningful phenotypes confirmed by a cardiologist. Using phenotypes extracted by \mname, simple logistic regression can achieve comparable predictive accuracy with deep learning techniques such as RNNs.%\kp{as in the abstract, i think we should state the advantage as well (e.g., comparable to RNN, but leading to a model which is much easier to interpret and debug)}  %, and phenotypes also have predictive capacity to discriminate cases from controls. %\kp{let's emphasize the importance of the HF subtyping problem here: e.g., saying that Precise HF subtyping is an unsolved clinical problem. (we could also cite a paper here)}
%\sy{we cannot say HF phenotyping is an unsolved clinical problem, there are few published studies already in HF phenotypes. it is fair to say "HF is a heterogeneous disease with complex mechanisms of cardiac deterioration over time, and once diagnosis, the response to treatment varies significantly, all of which implies existence of different phenotypes."}

\end{itemize}

%% file: 02-related_work.tex
\section{\textbf{Background \& Related Work}}
%\js{only one subsection here looks weird, probably need to add more about phenotyping with tensors}
%\kp{I feel you could save some space by omitting the following paragraph completely and start by describing the mode of the tensor.}
%\sy{Only talk about PARAFAC2 is not sufficient since those two methods are similar. Should talk about tenor model in general, and then specify several key progress in tensor factorization. Please use the presentation you did to analysts at Sutter}
%In this Section, we provide the necessary background for tensor operations. Then we give the complexity of some well known linear algebraic operations which facilitate the complexity comparison of different algorithms. Finally, we briefly illustrate the related work including: classic method for PARAFAC2 \cite{kiers1999parafac2}, original method for a coupled matrix and tensor factorization (CMTF) \cite{acar2011all} and the recent approach for Non-negative PARAFAC2 \cite{cohen2018nonnegative} and its complexity analysis.
 Table \ref{symbol} summarizes the notations used in this paper.
 \begin{table}[!ht] 
     \centering
      \caption{  % \kp{notation $Y_r$ is confusing here: why do we use subscript to denote column, it should just be Y(:, r) similar to row selection. this is also problematic if you compare with tensor slice indexing. i suggest removing that completely and changing all notation where we index columns as $Y(:, r)$}
      Notations
      }
      \label{symbol}
 \scalebox{0.75}{
 \begin{tabular}{ c|c}
 \hline
 \parbox[t]{1cm}{\centering Symbol} &  \parbox[t]{4cm}{ \centering Definition }  \\
 \hline
 *  & Element-wise Multiplication  \\
 $\odot$ & Khatri Rao Product \\
 \hline
 $\M{Y},\V{y}$ &  matrix, vector \\
 $\V{Y(i,:)}$ & the $i$-th row of $\M{Y}$ \\
 $\M{Y(:,r)}$ & the $r$-th column of $\M{Y}$ \\
 $\M{Y(i,r)}$ & element (i,r) of $\M{Y}$ \\
 $\M{X_k}$ &  Feature matrix of patient $k$ \\
 %$diag(\V{y})$ & Diagonal matrix with vector $\V{y}$ on diagonal \\
 $\text{diag}(\M{Y})$ & Extract the diagonal of matrix $\M{Y}$ \\
 $\text{vec}(\M{Y})$ & Vectorizing matrix $\M{Y}$ \\
 $\text{svd}(\M{Y})$ & Singular value decomposition on $\M{Y}$ \\
 $||\cdot||_F^2$ & Frobenius Norm \\
 $\text{max}(0,\M{Y})$ & max operator replaces negative values in \M{Y} with 0  \\
 $\M{Y} \geq0$ & All elements in \M{Y} are non-negative\\
 %$A, I, C, Q, Z$ & Matrices  \\

 %$\gamma$, $\lambda$ & Vector  \\
 %$\gamma_n, \lambda_n$ & Scalar \\
 \hline
 \end{tabular}}
 \end{table}

\subsection{\textbf{PARAFAC2 Model}}
%\kp{1) let's introduce the model with orthogonality constraints and state that those constraints are included to promote uniqueness 2) introduce uniqueness directly in the context of parafac2}
%\kp{maybe let's rephrase the title to simply PARAFAC2 and add one sentence at the end of this paragraph to mention SPARTan as a method to fit this model for sparse datasets, as well as COPA that adds constraints such as temporal smoothness to the model factors. the related work needs to be complete even if we mentioned those in the intro.}

The PARAFAC2 model \cite{kiers1999parafac2}, has the following objective function:
\begin{equation}
\small
\begin{aligned}
& \underset{\{\M{U_k}\}, \{\M{S_k}\}, \M{V}}{\text{minimize}}
& & \sum_{k=1}^{K} \frac{1}{2}||\M{X_k} -\M{U_k}\M{S_k}\M{V^T}||_F^2 \\
& \text{subject to}
& & \M{U_k}=\M{Q_k} \M{H}, \quad \M{Q_k^T} \M{Q_k}=\M{I}, 
\end{aligned}
\label{Classic_PARAFAC2_obj_func}
\end{equation}
where  ${\small\M{X_k}\in \mathbb{R}^{I_k \times J}}$ is the input matrix,  factor matrix {\small$\M{U_k} \in \mathbb{R}^{I_k \times R}$}, diagonal matrix {\small$\M{S_k} \in \mathbb{R}^{R \times R}$} , and factor matrix {\small$\M{V} \in \mathbb{R}^{J \times R}$} are output matrices.  Factor matrix {\small$\M{Q_k} \in \mathbb{R}^{I_k \times R}$} is an orthogonal matrix, and  {\small$\M{H} \in \mathbb{R}^{R \times R}$} where $\M{U_k}=\M{Q_k}\M{H}$.  SPARTan \cite{Perros2017-dh} is a method to fit this model for sparse datasets, as well as COPA \cite{afshar2018copa} incorporates different constraints such as temporal smoothness and sparsity to the model factors to produce more meaningful results. 
%\subsubsection{\textbf{Uniqueness}}~\label{sec:back_uniq}%: cross-product invariance constraint}
%\ho{I think uniqueness as a title is sufficient for subsection}

\noindent{\bf Uniqueness} property ensures that a decomposition is pursuing the true latent factors, rather than an arbitrary rotation of them. The unconstrained version of PARAFAC2 in~\eqref{Classic_PARAFAC2_obj_func} (without constraints \M{U_k}=\M{Q_k} \M{H} and \M{Q_k^T} \M{Q_k}=\M{I} ) is not unique. Assume \M{B} as an invertible $R \times R$ matrix and $\{\M{Z_k}\}$ as $R \times R$ diagonal matrices. Then, we can transform $\M{U_k}\M{S_k}\M{V^T}$ as:
\[
\M{U_k}\M{S_k}\M{V^T}=\underbrace{(\M{U_k}\M{S_k}\M{B^{-1}}\M{Z_k^{-1}})}_{\M{G_k}}\M{Z_k}\underbrace{(\M{B}\M{V^T})}_{\M{E^T}}%=\M{G_k}\M{Z_k}\M{E^T}
\]
which is another valid solution achieving the same approximation error~\cite{kiers1999parafac2} - this is problematic in terms of the interpretability of the result. To promote uniqueness, Harshman \cite{Hars1972b} introduced the \textbf{cross-product invariance constraint}, which dictates that $\M{U_k^T}\M{U_k}$ should be constant $\forall k\in\{1, \dots, K\}$. %\js{why do we need to cite this \cite{kiers1999parafac2}? remove?}
To achieve that, the following constraint is added: $\M{U_k}=\M{Q_k}\M{H}$ where $\M{Q}_k^T \M{Q}_k = \M{I}$, so that: $\M{U_k^T}\M{U_k}=\M{H^T}\M{Q_k^T}\M{Q_k}\M{H}=\M{H^T}\M{H}=\Phi$. 

\subsection{Unsupervised Computational Phenotyping}
A wide range of approaches applies tensor factorization techniques to extract phenotypes. \cite{ho2014limestone,ho2014marble,henderson2017granite,Perros2018kdd, zhao2019detecting, zhao2019using, jiang2019combining} incorporate various constraints (e.g., sparsity, non-negativity, integer) into regular tensor factorization to produce more clinically-meaningful phenotypes. \cite{Perros2017-dh,afshar2018copa} identify phenotypes and their temporal trends by using irregular tensor factorization based on PARAFAC2~\cite{Hars1972b}; yet, those approaches cannot model both dynamic and static features for meaningful phenotype extraction. As part of our experimental evaluation, we demonstrate that naively adjusting such PARAFAC2-based approaches to incorporate static information results in biased and less interpretable phenotypes. The authors of~\cite{YQCFP19aaai} proposed a collective non-negative tensor factorization for phenotyping purposes. However, the method is not able to jointly incorporate static information such as demographics with temporal features. Also they do not employ the orthogonality constraint on the temporal dimension, a strategy known to produce non-unique solutions~\cite{Hars1972b,kiers1999parafac2}.

%% file: 03-method.tex
\section{The \mname framework} 

\begin{comment}
Coupled Non-negative PARAFAC2 framework is a tool which decomposes an irregular tensor and a matrix where  factor matrix \M{S_k} is shared between them as shown in figure \ref{fig_Coupled_PARAFAC2}. %To the best of our knowledge this is the first time that someone tries to propose such a framework.
Enforcing the PARAFAC2 framework to handle both  temporal-evolving  and static information from the raw data  can improve the quality of clustering. For instance, in our main motivating application, phenotyping, we are combining the medical features in each clinic encounter with static information like age, race, gender, etc to improve the quality of phenotypes. %\ho{I think it would be helpful to spend 2-3 sentences re-motivating the need for coupled formulation and how it relates to phenotyping as an example}
Therefore, in this section we will formulate the general framework for Coupled Non-negative PARAFAC2 (\mname).
\end{comment}
\begin{comment}

\js{We jump right into the math without any intuition. People will get lost and ignore the method section. It is very important to explain the technical part gently and gradually with sufficient application support.}
\js{we should structure the method section as
1) Input and Output
2) Objective function and challenges in solving that
3) High level approach summary
3.1) Solve temporal constraints, U, V
   3.1.1 U, Q, H   
   3.1.2 V
3.2) Solve constraints for static features W, F
}
\js{Introduce the input $X_k$, A with the context of the phenotyping application. }
\end{comment}
\subsection{Intuition} \label{sec_inputoutput}
We will explain the intuition of \mname in the context of phenotyping application. 

\noindent{\bf Input data}  include both temporal and static features for all $K$ patients:
\begin{itemize}
\item \textbf{Temporal features (\M{X_k});} For patient $k$, we record the medical features for different clinical visits in matrix $\M{X_k} \in \mathbb{R}^{I_k \times J}$ where $I_k$ is the number of clinical visits and $J$ is the total number of medical features. Note that $I_k$ can be different for different patients. 
\item \textbf{Static features (\M{A}):} The static features like 
%age group,\footnote{Although age is strictly speaking a dynamic feature, we consider the age static for phenotyping as the time window for phenotyping study is often much smaller (6 months to 1 year), where age change within the study period is negligible.} 
gender, race, body mass index (BMI), smoking status \footnote{Although BMI and smoking status can change over time, in our data set these values for each patient are constant over time.} are recorded in $\M{A} \in \mathbb{R}^{K \times P}$ where $K$ is the total number of patients and $P$ is the number of static features. In particular, $\M{A}(k,:)$ is the static features for $k^{th}$ patient.
\end{itemize}

\noindent{\bf Phenotyping process} maps input data into a set of phenotypes, which involves the definition of phenotypes and temporal evolution.
 Figure \ref{fig_Coupled_PARAFAC2} illustrates 
the following model interpretation. 
First, \noindent{\it phenotype definitions} are shared by factor matrices \M{V} and \M{F} for temporal and static features, respectively. In particular, \textbf{\M{V(:,r)}} or \textbf{\M{F(:,r)}} are the $r^{th}$ column of factor matrix \M{V} or \M{F} which indicates the participation of temporal or static features in the $r^{th}$ phenotype.
Second, \noindent{\it personalized phenotype scores} for patient $k$ are provided in the diagonal matrix \M{S_k} where its diagonal element \M{S_k(r, r)} indicates the overall importance of the $r^{th}$ phenotype for patient $k$.
Finally, {\it temporal phenotype evolution} for patient $k$ is specified in factor matrix \M{U_k} where its $r^{th}$ column \M{U_k(:,r)} indicates the temporal evolution of  phenotype $r$ over all clinical visits of patient $k$.
%\js{we should consider adding a concrete phenotype example to illustrate what \textbf{\M{V(:,r)}}, \textbf{\M{F(:,r)}},  \M{S_k(r, r)} and \M{U_k} will be}
%\textbf{\M{F(:,r)}}, the $r^{th}$ column of factor matrix \M{F} indicates the participation of $P$ static features in the $r^{th}$ phenotype.
%\js{Make sure we use dynamic instead of temporal in all places}
% \begin{itemize}[leftmargin=*]
% \item \textbf{\M{U_k(:,r)}},  the $r^{th}$ column of factor matrix \M{U_k} indicates the temporal evolution of  phenotype $r$ over all clinical visits of patient $k$. 
% \item  \M{S_k} is a diagonal matrix where its diagnal element \M{S_k(r, r)} indicates the overall importance of the $r^{th}$ phenotype for patient $k$.
% \item \textbf{\M{V(:,r)}}, the $r^{th}$ column of factor matrix \M{V} indicates the participation of dynamic features in the $r^{th}$ phenotype for all patients.
% \item \textbf{\M{F(:,r)}}, the $r^{th}$ column of factor matrix \M{F} indicates the participation of $P$ static features in the $r^{th}$ phenotype.
% \end{itemize}
% The mixture of $r^{th}$ column of factor matrices \M{V} and \M{F} defines the phenotypes in our motivating example.
\subsection{\textbf{Objective function and challenges}}
We introduce the following objective:
\begin{equation}
\footnotesize
\begin{aligned}
& \underset{\substack{\{\M{U_k}\}, \{\M{Q_k}\}, \\ \M{H}, \{\M{S_k}\}, \M{V}, \M{F}}}{\text{minimize}}
& & \underbrace{\sum_{k=1}^{K}\Big( \frac{1}{2}||\M{X_k} -\M{U_k}\M{S_k}\M{V^T}||_F^2}_\text{PARAFAC2 (1) }\Big) +
 \underbrace{\frac{\lambda}{2}||\M{A}-\M{W}\M{F^T}||_F^2}_\text{Coupled Matrix (2)} +\underbrace{\sum_{k=1}^{K} \Big(\frac{\mu_k}{2}||\M{U_k}-\M{Q_k}\M{H}||_F^2}_\text{Uniqueness (3)}\Big) \\ 
& \text{subject to}
&  &  \M{Q_k^T} \M{Q_k}=\M{I}, \quad \M{U_k}\geq0, \quad \M{S_k }\geq0, \quad \text{for all k=1,...,K}  \\
%& &  &  \text{for all k=1,...,K} \\
& & & \V{W(k,:)}=\text{diag}(\M{S_k})  \quad \text{for all k=1,...,K} \\
& & &  \M{V }\geq0,  \quad \M{F }\geq0 \\
\end{aligned}
\label{NN_coupled_PARAFAC2}
\end{equation}
%\kp{it may be confusing at first sight that we claim that $\M{S_k}$ is shared but there is another factor $\M{W}$ for that. maybe we can color those two factors with the same color? and use the same color in the description below?}
\begin{comment}
\begin{equation}
\small
\begin{aligned}
& \underset{\{\M{U_k}\}, \{\M{S_k}\}, \M{V}, \M{F}}{\text{minimize}}
& & \sum_{k=1}^{K} \frac{1}{2}||\M{X_k} -\M{U_k}\M{S_k}\M{V^T}||_F^2 +\frac{\lambda}{2}||\M{A}-\M{W}\M{F^T}||_F^2   \\
& \text{subject to}
& & \M{U_k}=\M{Q_k} \M{H}, \quad \M{Q_k^T} \M{Q_k}=\M{I},   \\
& & & \M{U_k}\geq0, \quad \M{S_k}\geq0, \quad \M{V}\geq0, \quad \M{F }\geq0 
\end{aligned}
\label{NN_coupled_PARAFAC2_obj_func}
\end{equation}
\end{comment}
Objective function has three main parts as follows:
\begin{enumerate}[leftmargin=*]
\item The first part is related to fitting a PARAFAC2 model that factorizes a set of temporal feature matrices $\M{X_k}  \in \mathbb{R}^{I_k \times J}$ into {\small$\M{U_k} \in \mathbb{R}^{I_k \times R}$}, diagonal matrix {\small$\M{S_k} \in \mathbb{R}^{R \times R}$}, and {\small$\M{V} \in \mathbb{R}^{J \times R}$}.

\item The second part is for optimizing the static feature matrix \M{A} where ${\small\M{A}\in \mathbb{R}^{K \times P}}$, {\small$\M{W} \in \mathbb{R}^{K \times R}$}  and {\small$\M{F} \in \mathbb{R}^{P \times R}$}. $\lambda$ also is the weight parameter. Common factor matrices $\{\M{S_k}\}$ are shared between  static and temporal features by setting \V{W(k,:)}=diag(\M{S_k}).

\item %Instead forcing $\M{U}_k$ to be $\M{U_k}=\M{Q_k}\M{H}$ and  $\M{U_k}$ to be non-negative, since $\M{Q_k}$ may contain negative values. Therefore, 
The third part  enforces both non-negativity of the $\M{U_k}$ factor and also minimizes its difference to $\M{Q_k}\M{H}$. Due to the constraint $\M{Q}_k^T \M{Q}_k = I$, minimizing $||\M{U}_k - \M{Q}_k H||_F^2$ is encouraging that $\M{U_k^T}\M{U_k}$ is constant over K subjects, which is a desirable property from PARAFAC2 that promotes uniqueness, and thus enhancing interpretability~\cite{kiers1999parafac2}.

\item $\lambda$ and $\mu_k$ are weighting parameters which all set by user. For simplicity, we set $\mu_1=\mu_2=\cdots=\mu_K=\mu$.  %\js{how to set them, especially all different $\mu_k$. Is $\mu_k$ learned from the data or set by user?}%\ho{Maybe mention again the importance of uniqueness with respect to interpretability}%Penalty parameter $\mu_k$  controls the distance between the two terms. 
\end{enumerate}
% The first line of constraints promotes that if \M{U_k} be close to \M{Q_k}\M{H} then \M{U_k^T}\M{U_k} is constant over the $K$ matrices and equals \M{H^T}\M{H}. The second and third lines enforce non-negativity constraints to all the output factor matrices ($\{\M{U_k}\}, \M{V}, \{\M{S_k}\}, \M{F}$). 
The challenge in solving the above optimization problem lies in: 1)  addressing all the non-negative constraints especially on $\M{U_k}$, 2) trying to make \M{U_k^T}\M{U_k}  constant over $K$ subjects by making non-negative \M{U_k} as close as possible to \M{Q_k}\M{H} while \M{Q_k}\M{H} can contain negative values, and 3) estimating all factor matrices in order to best approximate both temporal and static input matrices. %2) developing a computationally efficient method for sparse input which is prevalent in phenotyping, our motivation application; %\kp{maybe omit this paragraph? this is the model section and we start talking about how to solve that. you could potentially integrate that with the start of the next section.}
%\kp{1) refers to sparse input and the fact that we need to develop sth comp. efficient for that. But, did we do anything to exploit or handle sparsity in this work? it seems that there is no novelty related to that (seems that we just use SPARTan to exploit sparsity), so why emphasize that here in the core method description? i think we may focus on stating that due to the non-convex nature of the problem, we resort to alternating optimization to solve that (solving for a certain block of parameters, by fixing the rest of them). Then, we should find a good way of motivating why AO-ADMM is the right approach of performing alternating optimization for this problem. why not pure-ALS?}
 %including the variable mode (\{\M{U_K}\}).
% \kp{Are we missing a constraint in~\eqref{NN_coupled_PARAFAC2_obj_func}? I had the impression that $\M{F}$ was also non-negative.}
%The Lagrangian for objective function \ref{NN_coupled_PARAFAC2_obj_func} is:
%\kp{Is the Lagrangian a new optimization problem? I was expecting to read a function where you would incorporate the constraints by introducing the Lagrange multipliers}
%\kp{There needs to be a header at this point indicating that we are describing the solution for $\M{Q}_k$ - I suggest you do the same for all the factor matrices so that the reader does not get lost}
%\js{we should explain what the challenge of the solving the objective function and our high-level approach, before we dive in and describe the details.}
%\\The objective function needs to address the following challenges:
\subsection{\textbf{Algorithm}}

%\kp{In the paragraph below, I think we should emphasize more the methodological novelty / mathematical re-formulations we do. The paragraph below bluntly states what we do, as if it's known beforehand how to solve and optimize all these sub-problems. Also, we over-emphasize the block-principal pivoting method, while our novelty is in how do you transform the original problem into simpler sub-problems which can then exploit the power of BPP and in general can be solved through NNLS. Overall, I would emphasize more the fact that we both frame each one of the sub-problems so that we can exploit the Block Coordinate Descent optimization framework, as well as exploit structure in the underlying computations (e.g., involving Khatri-Rao products) so that each one of the sub-problems is solved efficiently. Ari, please double-check whether this may be an accurate description, but in my understanding there is a lot of method novelty hidden here, which we should somehow concisely summarize and advertise (both in the paragraph below, and in Abstract + Intro).}

To optimize the objective function (\ref{NN_coupled_PARAFAC2}), we need to update $\{\M{Q_k}\}, \M{H}, \{\M{U_k}\}, \M{V},$ $\{\M{S_k}\},$ and \M{F} iteratively.  Although the original problem in Equation~\ref{NN_coupled_PARAFAC2} is non-convex, our algorithm utilizes the Block Coordinate Descent framework \cite{kim2014algorithms} to mathematically  reformulate the objective function (\ref{NN_coupled_PARAFAC2}) into  simpler sub-problems. %\kp{maybe we cite the Block Coordinate Descent framework here}
In each iteration, we update $\{\M{Q_k}\}$ based on Orthogonal Procrustes problem \cite{schonemann1966generalized} which ensures an orthogonal solution for each \M{Q_k}.  Factor matrix \M{H} can be solved efficiently by  least square solvers. For factor matrices $\{\M{U_k}\}, \M{V}, \{\M{S_k}\}, \M{F}$ we reformulate objective function (\ref{NN_coupled_PARAFAC2}) so that  the factor matrices are instances of the non-negativity constrained least squares (NNLS) problem. 
Each NNLS sub-problem  is a convex problem and the optimal solution can be found easily. We use {\it block  principal pivoting method} \cite{kim2011fast}  to solve each NNLS sub-problem.   %this process monotonically reduces the objective function. %\js{it will be important to add convergence proof in the supplement}
The block principal pivoting method achieved state-of-the-art performance on NNLS problems compared to other optimization techniques \cite{kim2011fast}.
%The block  principal pivoting method in \cite{kim2011fast} demonstrated the state of the art performance comparing to other optimization techniques.
We provide the details about NNLS problems in the supplementary material section. We also exploit structure in the underlying computations (e.g., involving Khatri-Rao products) so that each one of the sub-problems is solved efficiently.  Next, we summarize the solution for each factor matrix. %\ari{add some intuition on why BPP is good.} %For all but the $\{\M{Q_k}\}$ factor matrices, AO-ADMM~\cite{huang2016flexible} enables us to efficiently handle a wide range of constraints on factor matrices including  the non-negativity constraint by performing element-wise operations. We resort to a Procrustes-based solution for $\{\M{Q_k}\}$ due to the orthogonality constraint on this set of factors. %Computational caching (performing Cholesky decomposition) also ensure that each update performs efficiently.  
%In this part, we provide the details for updating each factor matrix.
%\kp{is it really just ADMM being used here? it does not seem so - for example after going through the description of $\M{Q}_k$ factor's solution, it seems that we solve it in an ALS style. i think we be more precise here in terms of what is really the methodology that we are using and the intuition behind why we think this is the right way to solve the problem} 
\subsubsection{\textbf{Solution for factor matrix $\M{Q_k}$}}
We can reformulate  objective function \ref{NN_coupled_PARAFAC2}  with respect to $\M{Q_k}$ as follows :

\begin{equation}
\small
\begin{aligned}
& \underset{\M{Q_k}}{\text{minimize}}
& & \mu_k\norm{  \M{U_k} \M{H}^T-\M{Q_k}}_F^2 \\
& \text{subject to}
& &  \M{Q_k^T} \M{Q_k}=\M{I}
\end{aligned}
\label{Equation_Q_k}
\end{equation}

 More mathematical details about converting Equation \ref{NN_coupled_PARAFAC2} to \ref{Equation_Q_k}  are provided in supplementary material section. The optimal value of $\M{Q_k}$ can be computed via Orthogonal Procrustes problem \cite{schonemann1966generalized} which has the closed form solution $\M{Q_k}=\M{B_k}\M{C_k}^T$ where $\M{B_k} \in R^{I_k \times R}$ and $\M{C_k} \in R^{R \times R}$ are the right and left singular vectors of $ \mu_k \M{U_k} \M{H}^T$. Note that each $\M{Q_k}$ can contain negative values.

\begin{comment}
After removing the constant terms and applying $tr(\M{A})=tr(\M{A}^T)$, we have $tr((\M{H}\M{S_k}\M{V}^T\M{X_k}^T+\mu_k \M{H}\M{U_k}^T)\M{Q_k})$ which equals to:
\begin{equation}
\small
\begin{aligned}
& \underset{\M{Q_k}}{\text{minimize}}
& &  \frac{1}{2}||\M{X_k} \M{V} \M{S_k} \M{H^T} + \mu_k \M{U_k} \M{H}^T-\M{Q_k}||_F^2 \\
& \text{subject to}
&  &  \M{Q_k^T} \M{Q_k}=\M{I}
\end{aligned}
\label{rewrite_Q}
\end{equation}
The optimal value of $\M{Q_k}$ can be computed via Orthogonal Procrustes problem \cite{schonemann1966generalized} which has the closed form solution $\M{Q_k}=\M{B_k^T}\M{C_k}$ where $\M{B_k} \in R^{I_k \times R}$ and $\M{C_k} \in R^{R \times R}$ are the right and left singular vectors of $\M{X_k} \M{V} \M{S_k} \M{H^T} + \mu_k \M{U_k} \M{H}^T$.
\\After updating factor matrix $\{\M{Q_k}\}$, we use the AO-ADMM algorithm to update the remaining factors which is efficient to handle non-negativity constraint by performing element-wise operations. %\kp{what would be the problem of ALS (is it just easier to implement non-negativity constraint?), this keeps coming up as a thought. we need to be able to persuade the reader that what we propose makes sense intuitively and this is what they should do}
\end{comment}
\subsubsection{\textbf{Solution for factor matrix $\M{H}$}} The objective function with respect to \M{H} can be rewritten as: 

\begin{equation}
\small
\begin{aligned}
& \underset{\M{H}}{\text{minimize}}
& & \sum_{k=1}^{K} \frac{\mu_k}{2}\norm{\M{Q_k}^T\M{U_k}-\M{H}}_F^2
\end{aligned}
\label{H_equationmain}
\end{equation}
\begin{comment}
\begin{equation}
\small
\begin{aligned}
& \underset{\M{H}}{\text{minimize}}
& & \frac{1}{2}\norm{  \begin{bmatrix}
\sqrt{\mu_1}\M{Q_1} \\
\sqrt{\mu_2}\M{Q_2}  \\
. \\
\sqrt{\mu_K}\M{Q_K}
\end{bmatrix} \M{H}-\begin{bmatrix}
\sqrt{\mu_1}\M{U_1} \\
\sqrt{\mu_2}\M{U_2}  \\
. \\
\sqrt{\mu_K}\M{U_K}
\end{bmatrix}}_F^2 \\
\end{aligned}
\label{H_equation}
\end{equation}
\end{comment}
without any constraint. The details about updating factor matrix \M{H} are provided in supplementary material section.  The value of \M{H} that minimizes Equation \ref{H_equationmain} also minimizes Equation \ref{NN_coupled_PARAFAC2}. The update rule for factor matrix \M{H} has the following form:
\[
\M{H}=\frac{\sum_{k=1}^{K} \mu_k \M{Q_k}^T\M{U_k}}{\sum_{k=1}^{K} \mu_k}.
\]
\subsubsection{\textbf{Solution for phenotype evolution matrix $\M{U_k}$}} After updating the factor matrices \M{Q_k}, \M{H}, we focus on solving for \M{U_k}. In classic PARAFAC2~\cite{Hars1972b,kiers1999parafac2}, this factor is retrieved through the simple multiplication  $\M{U}_k =\M{Q_k}\M{H}$. However, due to interpretability concern, we prefer temporal factor matrix $\M{U}_k$ to be non-negative because the temporal phenotype evolution for patient k (\M{U_k}) should never be negative. As shown in the next section, a naive enforcement of non-negativity ($max(0, \M{Q}_k~\M{H})$) violates the important uniqueness property of PARAFAC2 (the model constraints). Therefore, we consider $\M{U_k}$ as an additional factor matrix, constrain it to be non-negative and minimize its difference to $\M{Q}_k\M{H}$.

%Both \M{Q_k} and \M{H} may have negative values, however, we want to make $\{\M{U_k}\}$ to be non-negative and also as close as possible to $\M{Q_k}\M{H}$. \kp{this is not clear. the reader will think "why do you even need to do that since you can just multiply those two terms and get the factor?" we really need to add more intuition (again) on why we do need this additional constraint}

%\kp{one way to motivate the need for this additional constraint could be the following: Due to the non-negativity of the input data, we prefer all the output factors to be non-negative so that interpretability is enhanced. As shown in the experiments, a naive enforcement of non-negativity ($max(0, \M{Q}_k~\M{H})$) violates important properties of PARAFAC2 which promote its uniqueness - this is why we introduce this constraint.}
The objective function with respect to $\M{U_k}$ can be combined into the following NNLS form:
%\ari{fic the notaion}
\begin{equation}
\small
\begin{aligned}
& \underset{\M{U_k}}{\text{minimize}}
& & \frac{1}{2}\norm{ \begin{bmatrix}
\M{V}\M{S_k} \\
\sqrt{\mu_k} \M{I}
\end{bmatrix} \M{U_k}^T-\begin{bmatrix}
\M{X_k}^T \\
\sqrt{\mu_k} \M{H}^T\M{Q_k}^T
\end{bmatrix}}_F^2 \\
& \text{subject to}
& &  \M{U_k} \ge 0
\end{aligned}
\label{Equation_U_k}
\end{equation}
As we mentioned earlier, we update factor matrix \M{U_k} based on block principal pivoting method. %\js{need a sentence about how to solve this objective}

\begin{comment}
We define an auxiliary variable $\M{\overbar{U_k}}$ to handle the non-negativity constraint on \M{U_k} where $\M{U_k}=\M{\overbar{U_k}}$. The update rules for each factor matrix $\M{U_k}$ by using ADMM algorithm have the following iterations: %\kp{similar to my comment above, i think the description lacks intuition}
\begin{equation}
\small
\begin{aligned}
\M{U_k} &:= \frac{(\mu_1 \M{Q_k}\M{H}+ \rho (\M{\overbar{U_k}}+\M{D}))}{\mu_k+\rho}\\
%\overbar{\M{U_k}}& :=\arg\min_{\overbar{\M{U_k}}} c(\overbar{\M{U_k}}) +\frac{\rho}{2} ||\M{U_k}-\M{\overbar{U_k}}+\M{D}||_F^2 \\
\overbar{\M{U_k}}& := \arg\min_{\overbar{\M{U_k}}} n(\overbar{\M{U_k}}) +\frac{\rho}{2} ||\overbar{\M{U_k}}-\M{U_k^T}+\M{D}_{U_k^T}||_F^2\\
& =max(0,\M{U_k}-\M{D}) \\
\M{D} &:=\M{D}+\M{U_k}-\M{\overbar{U_k}}
\end{aligned}
\label{update_U_k}
\end{equation}
\end{comment}
\subsubsection{\textbf{Solution for temporal phenotype definition $\M{V}$ }}  Factor matrix \M{V} defines the participation of temporal features in different phenotypes. By considering Equation \ref{NN_coupled_PARAFAC2}, factor matrix \M{V} participates in the PARAFAC2 part with non-negativity constraint.  Therefore, the objective function for factor matrix \M{V} have the following form:

\begin{equation}
\small
\begin{aligned}
& \underset{\M{V}}{\text{minimize}}
& & \frac{1}{2}\norm{  \begin{bmatrix}
\M{U_1} \M{S_1} \\
\M{U_2} \M{S_2} \\
. \\
\M{U_K} \M{S_K}
\end{bmatrix} \M{V}^T-\begin{bmatrix}
\M{X}_1 \\
\M{X}_2  \\
. \\
\M{X}_K
\end{bmatrix}}_F^2 \\
& \text{subject to}
& &  \M{V} \ge 0
\end{aligned}
\label{V_equation}
\end{equation}

In order to update $\M{V}$ based on block principal pivoting, we need to calculate $(\M{U_k}\M{S_k})^T (\M{U_k}\M{S_k})$ and $\M{U_k}\M{S_k}\M{X_k}$  for all $K$ samples which can be calculated in parallel.
\subsubsection{\textbf{Solution for factor matrix $\M{W}$ or $\{\M{S}_k$\}}} The objective function with respect to $\M{W}$ yields the following format:

\begin{equation}
\small
\begin{aligned}
& \underset{\M{S_k}}{\text{minimize}}
& & \sum_{k=1}^{K}\Big( \frac{1}{2}||\M{X_k} -\M{U_k}\M{S_k}\M{V^T}||_F^2\Big) +\frac{\lambda}{2}||\M{A}-\M{W}\M{F^T}||_F^2  \\
& \text{subject to}
& &  \M{S_k }\geq0 \\
& & & \V{W(k,:)}=\text{diag}(\M{S_k})  \quad \text{for all k=1,...,K} \\
\end{aligned}
\label{Equation_W}
\end{equation}

As we mentioned earlier, factor matrices $\{\M{S_k}\}$ is shared between PARAFAC2 input and matrix \M{A} where \V{W(k,:)}=diag(\M{S_k}). By knowing $\text{vec}(\M{U_k} \M{S_k} \M{V}^T)=(V \odot \M{U_k})\V{W(k,:)}^T$, Equation~\ref{Equation_W} can be rewritten as:

\begin{equation}
\small
\begin{aligned}
& \underset{\M{S_k}}{\text{minimize}}
& & \frac{1}{2}\norm{\begin{bmatrix}
\M{V} \odot \M{U_k}\\
\sqrt{\lambda}\M{F}  
\end{bmatrix} \M{W(k,:) }^T- \begin{bmatrix}
\text{vec}(\M{X_k}) \\
\sqrt{\lambda} \M{A(k,:)}^T 
\end{bmatrix} }_F^2 \\
& \text{subject to}
& &  \M{W(k,:) }\geq0 
\end{aligned}
\label{S_k_equation}
\end{equation}
where  $\odot$ denotes Khatri-Rao product.  We can solve the rows of factor matrix \M{W} (\V{W(k,:)} or diag(\M{S_k})) separately and in parallel. To update each factor matrix $\M{S_k}$ we need to compute two time-consuming operations: 1)$(\M{V} \odot \M{U_k})^T (\M{V} \odot \M{U_k})$ and 2)$(\M{V} \odot \M{U_k})^T \text{vec}(\M{X_k})$. The first operation can be replaced with $\M{V}^T\M{V} * \M{U_k}^T \M{U_k}$ where * denotes element-wise(hadamard) product \cite{van2000ubiquitous}. The second operation also can be replaced with $diag(\M{U_k} \M{X_k} \M{V}^T)$ \cite{van2000ubiquitous}.  Therefore, the time-consuming Khatri-Rao Product doesn't need to be explicitly formed. Each row of W can be efficiently updated via block principal pivoting. 

%The second one is called MTTKRP operation. A naive implementation of MTTKRP operation needs huge storage and has  a time-consuming computational cost. Here, we are trying to make MTTKRP operation as efficient as possible by 1) making it parallel into $J$ blocks 2) taking advantage of sparsity structure in \M{X_k} to diminish the computations. The MTTKRP can be formulated as follows:
\begin{comment}
\begin{equation}
\small
(\M{V} \odot \M{U_k})^T vec(\M{X_k})=\sum_{\substack{j=1\\ \M{X_k}(j,:) \neq zeros}}^{J} (\M{V(j,;)}*\M{U_k})*\M{X_k}(:,j)=\sum_{\substack{j=1\\ \M{X_k}(j,:) \neq zeros}}^{J} \M{V(j,;)}*(\M{U_k}*\M{X_k}(:,j))
\label{MTTKRP_operation}
\end{equation}
\end{comment}
%\js{isn't this the same idea of column sparsity in SPARTAN? let's make sure and cite appropriately} Many columns of \M{X_k} contain zero values and for those cases we avoid multiplications. 

%where  $\M{L_W} \M{L_W}^T=chol\Big((\M{H^T} \M{H} * \M{V^T} \M{V})+ \lambda\M{F^T} \M{F}+ \rho \M{I}\Big)$ and $\M{L_W} \in \mathbb{R}^{R \times R} $. MTTKRP \cite{bader2007efficient}  operation is a bottleneck operation for sparse datasets. Therefore, for Equations \ref{update_H}, \ref{update_V}, and \ref{update_W} we use the fast MTTKRP calculation proposed in the SPARTan approach~\cite{Perros2017-dh}.

\subsubsection{\textbf{Solution for static phenotype definition $\M{F}$}} Finally, factor matrix \M{F} represents the participation of static features for the phenotypes. The objective function for factor matrix \M{F} have the following form:
\begin{equation}
\small
\begin{aligned}
& \underset{\M{F}}{\text{minimize}}
& & \frac{\lambda}{2}\norm{\M{W}\M{F^T}-\M{A}}_F^2 \\
& \text{subject to}
& &  \M{F} \ge 0
\end{aligned}
\label{Equation_F}
\end{equation}
which can be easily updated via block principal pivoting.

\subsection{Phenotype inference on new data} \label{proj_patient}
%\kp{let's be careful to adjust this section depending on whether we want to include the predictive model section eventually. if we don't include that, a good way to motivate this section may be that: In practice, if \mname is deployed in a hospital (e.g., for clinical decision support), we would like to be able to estimate the phenotype membership of new patients based on a fixed set of phenotype definitions, which have been clinically evaluated.}
%\kp{i think projecting a patient onto a fixed model is important by itself (e.g., finding similar patients for clinical decision support purposes) and does not need to be motivated through the additional predictive task. we can mention the prediction task at the end of the paragraph as additional motivation, but i think we should separate the discussion and focus on describing the methodology of projecting here. after all, we do not have anything novel to propose regarding the prediction itself.}

Given phenotype definition ($\M{V}, \M{F}$) and factor matrix \M{H} for some training set, \mname can project data of new unseen patients into the existing low-rank space.
This is useful because healthcare provider may want to fix the phenotype definition while score new patients with those existing definitions. Moreover, such a methodology enables using the low-rank representation of patients such as ($\M{S_k}$) as feature vectors for a predictive modeling task. 

%can be used for prediction tasks to help predict whether a new patient will be diagnosed with heart failure or not.
%Early detection of heart failure can reduce cost and delay the progression of it. \cite{choi2016using}. 
%because we may not always want to re-run the entire decomposition for new patients. Furthermore, after learning the low-rank phenotipic representation \mname is able to predict whether a new patient will be diagnosed with heart failure or not. %\kp{i think we should separate the description of projection + prediction. after all, one may be interested in solely projecting a patient and not predicting sth based on the low-rank representation.}%\ho{Maybe to address Kimis's point, you could start by the fact that you may not always want to re-run the entire decomposition for new patients, then maybe in that context it makes sense?}.
Suppose,  $\{\M{X_1}, \M{X_2}, ..., \M{X}_{N^{'}} \} $ represents the temporal information of unseen patients $\{1,2,..., N^{'}\}$  and $\M{A}^{'} \in \mathbb{R}^{N^{'} \times P}$ indicated their static information. \mname is able to project the new patient's information into the existing low-rank space (\M{H}, \M{V}, and \M{F}) by optimizing \{\M{Q_n}\}, $\{\M{U_n}\}$ and $\{\M{S_n}\}$ for the following objective function:
\begin{equation}
\footnotesize
\begin{aligned}
& \underset{\substack{\{\M{Q_n}\}, \{\M{U_n}\}, \\ \{\M{S_n}\}}}{\text{minimize}}
& & \sum_{n=1}^{N'}\Big( \frac{1}{2}||\M{X_n} -\M{U_n}\M{S_n}\M{V^T}||_F^2\Big) + \frac{\lambda}{2}||\M{A^{'}}-\M{W}\M{F^T}||_F^2 \\
& & &   + \sum_{n=1}^{N'} \Big(\frac{\mu_n}{2}||\M{U_n}-\M{Q_n}\M{H}||_F^2\Big) \\ 
& \text{subject to}
& &  \M{Q_n^T} \M{Q_n}=\M{I}, \quad \text{for all } n=1,...,N'  \\
& & & \M{U_n}\geq0, \quad \M{S_n }\geq0 \quad \text{for all } n=1,...,N' \\
\end{aligned}
\label{projection_equation}
\end{equation}

Updating factor matrices $\{\M{Q_n}\}$ is based on Equation \ref{Equation_Q_k}. $\{\M{U_n}\}$ can be minimized based on Equation \ref{Equation_U_k}. Finally,  \M{W} can be updated based on Equation \ref{S_k_equation} where $diag(\M{S_n})=W(n,:)$. Factor matrix \M{W} represents the personalized phenotype scores of all patients.

%% file: 04-experiments.tex
\section{Experimental Results}\label{sec:exp}
We focus on answering the following:
%\ari{Please double check if the questions are aligned with the experiment section.} \kp{write the following list in exact sequence with the sections and also refer to each section. e.g., 1) Does \mname recover the true factor matrices? How does enforcing uniqueness correlates with recovery in the presence of noise? (Section~\ref{recovery_true_fac}) }
%\kp{we need a really clear organization of the following enumeration and we really need each question to be connected to the corresponding section (ideally, sections should be sequentially organized according to the following). }
\begin{enumerate}
    %\item  Can we preserve the uniqueness property while achieving high accuracy?
    
    %\item Is \mname scalable enough to handle increasing number of patients.? %\kp{let's be specific because scalability may mean lots of diff things (e.g., scaling to larger number of compute nodes). let's say scalable enough to handle increasing amount of patients.}
    \item [Q1.] Does \mname preserve accuracy and the uniqueness-promoting constraint, while being fast to compute?  
     %\kp{i think we should rephrase as: how does the static information added in \mname would improve predictive performance in HF prediction? we should be really focused on our comparison because people may claim sth like "you did not even compare with RNN etc"}
    \item [Q2.] How does \mname scale for increasing number of patients ($K$)?
    \item [Q3.] Does the static information added in \mname improve predictive performance for detecting heart failure?
    \item [Q4.] Are the heart failure phenotypes produced by \mname meaningful to an expert cardiologist?
    %\item Does \mname recover the true factor matrices? How does promoting uniqueness correlate with recovery in the presence of noise? (In supplementary section)
\end{enumerate}

\subsection{\textbf{Data Set Description}}
Table~\ref{tab:summry_data} summarizes the statistics of  data sets.

\begin{table}[ht]
  \centering
  \caption{\footnotesize Summary statistics of two real  data sets.}
  \scalebox{0.75
  }{
    \begin{tabular}{cccccc}
    \multicolumn{1}{l}{Dataset} & \multicolumn{1}{l}{\# Patients} & \multicolumn{1}{l}{\# Temporal Features} & \multicolumn{1}{l}{Mean($I_k$)} & \multicolumn{1}{l}{\# Static Features} \\
    \midrule
    Sutter  & 59,480 & 1164  & 29   & 22  \\
    CMS   & 151,349 & 284   & 50  & 30  \\
    \hline
    %Case Study & 2516  & 662 &  25  & 28\\
    \end{tabular}%
    }
  \label{tab:summry_data}%
\end{table}%

\textbf{Sutter:}
%This dataset is from Sutter Palo Alto Medical Foundation, a  large primary care and multispecialty group practice.
%The dataset used for experiments is from a previously described predictive modeling study that included heart failure cases and group matched controls \cite{choi2016using}.
The data set contains the EHRs for patients with new onset of heart failure and matched controls (matched by encounter time, and age). It includes 5912 cases and 59300 controls. 
For all patients, encounter features (e.g., medication orders, diagnosis) were extracted from the electronic health records. We use standard medical concept groupers to convert the available ICD-9 or ICD-10 codes of diagnosis to Clinical Classification Software (CCS level 3) ~\cite{slee1978international}. We also group the normalized drug names based on unique therapeutic
%On all cases and controls, electronic health record data were extracted on encounters and encounter features (e.g., medication orders, diagnosis). We use standard medical concept groupers to convert the available ICD-9 or ICD-10 codes of diagnosis to Clinical Classification Software (CCS level 3) ~\cite{slee1978international}. We also group the normalized drug names based on unique therapeutic
sub-classes using the Anatomical Therapeutic Chemical (ATC) Classification System. Static information of patients includes their gender, age, race, smoking status, alcohol status and BMI. %The list of static features are provided in supplementary section. %We define 3 different age bands based on the first and third quartiles of the age sampling distribution: the bottom quartile (age$<$64), the middle two quartiles (64 $\leq$age $\leq$ 80), and the top quartile (80$<$age). %\kp{the second quartile is exactly the median, you don't seem to be using second quartile at all, but you rather define everything based on Q1 and Q3. let's rephrase if my understanding of what we do is correct}

\textbf{Centers for Medicare and Medicaid (CMS):}\footnote{\url{https://www.cms.gov/Research-Statistics-Data-and-Systems/Downloadable-Public-Use-Files/SynPUFs/DE_Syn_PUF.html}} The next data set is CMS 2008-2010 Data Entrepreneurs' Synthetic Public Use File (DE-SynPUF). The goal of CMS data set is to provide a set of realistic data by protecting the privacy of Medicare beneficiaries  by using 5\% of real data to synthetically construct the whole dataset.  We extract the ICD-9 diagnosis codes and convert them to CCS diagnostic categories as in the case of Sutter dataset.  %We use 30 static features including the race, gender, age bands and several chronic conditions  (diabetes, Alzheimer, depression, kidney disease).

\subsection{\textbf{Evaluation metrics:}}
%Decomposition accuracy is evaluated as an average of fit for PARAFAC2 tensor and matrix: $FIT= \frac{FIT_T+FIT_M}{2}$ where  $FIT_T= 1-\frac{\sum_{k=1}^{K} ||\M{X_k} -\M{U_k} \M{S_k}\M{V^T}||^2}{\sum_{k=1}^{K} ||\M{X_k}||^2}$ and $FIT_M= 1-\frac{ ||\M{A} -\M{W} \M{F^T}||^2}{ ||\M{A}||^2}$. FIT can be considered as the proportion of input data that can be explained by the model. 
%\kp{Please use the definition of RMSE I had shared with you, even if the one you mention is equivalent and you compute it that way. People will stare at this to make sure that this is indeed RMSE, and this is unnecessary burden for the reader.}
\begin{itemize}
    \item \textbf{RMSE: }
    Accuracy is evaluated as the Root Mean Square Error (RMSE) which is a  standard measure used in coupled matrix-tensor factorization literature~\cite{choi2017fast,beutel2014flexifact}. 
    % We define
    % \[ N:=\sum\limits_{k=1}^K \sum\limits_{i=1}^{I_k} \sum\limits_{j=1}^J 1\]
    % to be the total number of elements contained in our input collection of matrices $\M{X_k} \in \mathbb{R}^{I_k\times J}, \forall k=1,...,K$. We also define
    % \[ M:= \sum\limits_{k=1}^K \sum\limits_{j=1}^P 1 
    % \]
    % to be the total number of elements of the input matrix $\M{A} \in \mathbb{R}^{K\times P}$ containing static feature information. Then, RMSE can be defined as:
%\[
%RMSE=\sqrt{\frac{\sum_{k=1}^{K}\sum_{i=1}^{I_k \times J} %(x_i-\hat{x_i})^2+\sum_{i=1}^{K \times P} %(a_i-\hat{a_i})^2}{\sum_{k=1}^{K} (|\M{X_k}|)+|\M{A}|}}
%\]
 Given  input collection of matrices $\M{X_k} \in \mathbb{R}^{I_k\times J}, \forall k=1,...,K$ and static  input matrix $\M{A} \in \mathbb{R}^{K\times P}$, we define
\begin{equation}
%\tiny
\text{RMSE}=\sqrt{\frac{
\sum\limits_{k=1}^K \sum\limits_{i=1}^{I_k} \sum\limits_{j=1}^J
( \M{X_k}(i,j) - \M{\hat{X}_k}(i,j) )^2
+\frac{\lambda}{2}
\sum\limits_{k=1}^K \sum\limits_{j=1}^P
 (\M{A}(i,j) -  \M{\hat{A}}(i,j) )^2} 
 {\sum_{k=1}^K (I_k\times J) + K\times P}}
\end{equation}
%a true element in matrix \M{X_k} and element  $\hat{x_i}$ is the estimated one. Also  $a_i$ and $\hat{a_i}$  are also actual and estimated   values in  matrix \M{A}.  $|X_k|$  indicates the size of matrix \M{X_k} 
 %and $|\M{A}|$ represents the static matrix size including all zeros. %\js{not clear what the size of matrix means. Are you saying the number of elements including zero? We should clarify }Also $\hat{x_i}$ is the actual PARAFAC2 input element and $x_i$ is the estimated element. \js{usually, $\hat{x_i}$ is the estimate while $x_i$ is the actual. check and consider swap. The same comment applies to $a_i$ } 
 $\M{X_k}(i,j)$ denotes the $(i,j)$ element of input matrix $\M{X_k}$ and $\M{\hat{X}_k}(i,j)$ its approximation through a model's factors (the $(i,j)$ element of the product $\M{U_k}\M{S_k}\M{V^T}$ in the case of \mname). Similarly, $\M{A}(i,j)$ is the $(i,j)$ element of input matrix $\M{A}$ and $\M{\hat{A}}(i,j)$ is its approximation (in \mname, this is the $(i,j)$ element of $\M{W}\M{F}^T$).
 
\item \textbf{Cross-Product Invariance (CPI):}
We use CPI to assess the solution's uniqueness, since this is the core constraint promoting it~\cite{kiers1999parafac2}.
In particular we check whether $\M{U}_k^T \M{U}_k$ is close to constant ($\M{H^T}\M{H}$) $\forall k\in \{1, \dots, K\}$). The \textit{cross-product invariance} measure is defined as:
\[
\text{CPI}=1-\frac{\sum_{k=1}^{K} || \M{U_k^T}\M{U_k}- \M{H^T}\M{H}||^2_F}{\sum_{k=1}^{K} ||\M{H^T}\M{H}||^2_F}.
\]
\\The range of  cross-product invariance is between {\small$[-\infty,1]$}, with values close to 1 indicating  unique solutions ($\M{U}_k^T \M{U}_k$ is close to constant). %\js{check my change here} \kp{let's add citation to support this argument}}

\item \textbf{Area Under the ROC Curve (AUC):} %\kp{AUC description can be improved, currently it is kind of vague}
Examines classification model’s performance  when the data is imbalance by comparing the actual and estimated labels. %In our heart failure task the labels are whether a patient is diagnosed with HF or not.
We use AUC on the test set to evaluate predictive model performance. %\kp{check if we will keep that}

\end{itemize}

\subsection{\textbf{Q1. \mname is fast, accurate and preserves uniqueness-promoting constraints}} \label{sec:exp_fast_Acc}

\subsubsection{\textbf{Baseline Approaches:}}
In this section, we compare \mname with methods that incorporate non-negativity constraint on all factor matrices.  %including $\{\M{U_k}\}$.
Note that SPARTan \cite{Perros2017-dh} and COPA \cite{afshar2018copa} are not able to incorporate non-negativity constraint on factor matrices $\{\M{U_k}\}$.
%\ho{I think you can nuke the bullet point and just make it a single baseline approach unless your plan is to include SPARTan.}
%\js{what is our justification to not include SPARTan and COPA as baselines? probably need to state that to avoid criticism}
\begin{itemize}
    \item  \textbf{Cohen+~\cite{cohen2018nonnegative}:} Cohen et al. proposed a PARAFAC2 framework which imposes  non-negativity constraints on all factor matrices % including the varying mode (\M{U_k}) 
    based on non-negative least squares algorithm \cite{kim2014algorithms}. We modified this method to handle the situation where static matrix \M{A} is coupled with PARAFAC2 input based on Figure~\ref{fig_Coupled_PARAFAC2}. 
    %\js{did we extend the original Cohen to handle static features? If so, we should rephrase the previous sentence to state that clearly. If we extended Cohen, why not extending to SPARTAN and COPA and add those as baselines?(I'm not saying we should do that but we should have a reason about not doing that)}
    To do so, we add $\frac{\lambda}{2}||\M{A}-\M{W}\M{F^T}||_F^2$ to their objective function and updated both factor matrices \M{W} and \M{F} in an Alternating Least Squares manner, similar to how the rest of the factors are updated in~\cite{cohen2018nonnegative}.
%\js{if this is related to COPA, we should cite and mention COPA to avoid any confusion that we didn't compare to other known parafac2 methods. maybe we should call this COPA+} \kp{COPA+ is a really good idea}

\item \textbf{COPA+:}  One simple and fast way to enforce non-negativity constraint on factor matrix \M{U_k} is to compute $\M{U_k}$ as: $\M{U_k}:=max(0,\M{Q_k}\M{H})$, where $max()$ is taken element-wise to ensure non-negative results. Therefore, we modify the implementation in \cite{afshar2018copa} to handle both the PARAFAC2 input and the static matrix \M{A} and then apply the simple heuristic to make $\{\M{U_k}\}$ non-negative. %This heuristic approach can be viewed as a special case of our proposed method where the hyper-parameters $\mu_k, \forall k\in\{1, \dots, K\}$ are all set to zero, then the term minimizing the difference between $\M{U_k}$ and $\M{Q_k}\M{H}$ is inactive.
We will show in the experimental results section that how this heuristic method no longer guarantees unique solutions (violates model constraints). %\ho{I would reframe as this heuristic no longer guarantees unique solutions.}
\end{itemize}
 We provide the details about tuning hyperparameters in the supplementary material section.

\subsubsection{\textbf{Results:}} 
\begin{figure}[t]
\centering
    \begin{subfigure}[t]{0.4\textwidth}
        \includegraphics[height=0.25 in]{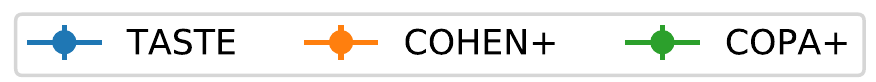}
         \label{fig:label}
    \end{subfigure}\\

    \begin{subfigure}[t]{0.37\textwidth}
        \includegraphics[height=1.4 in]{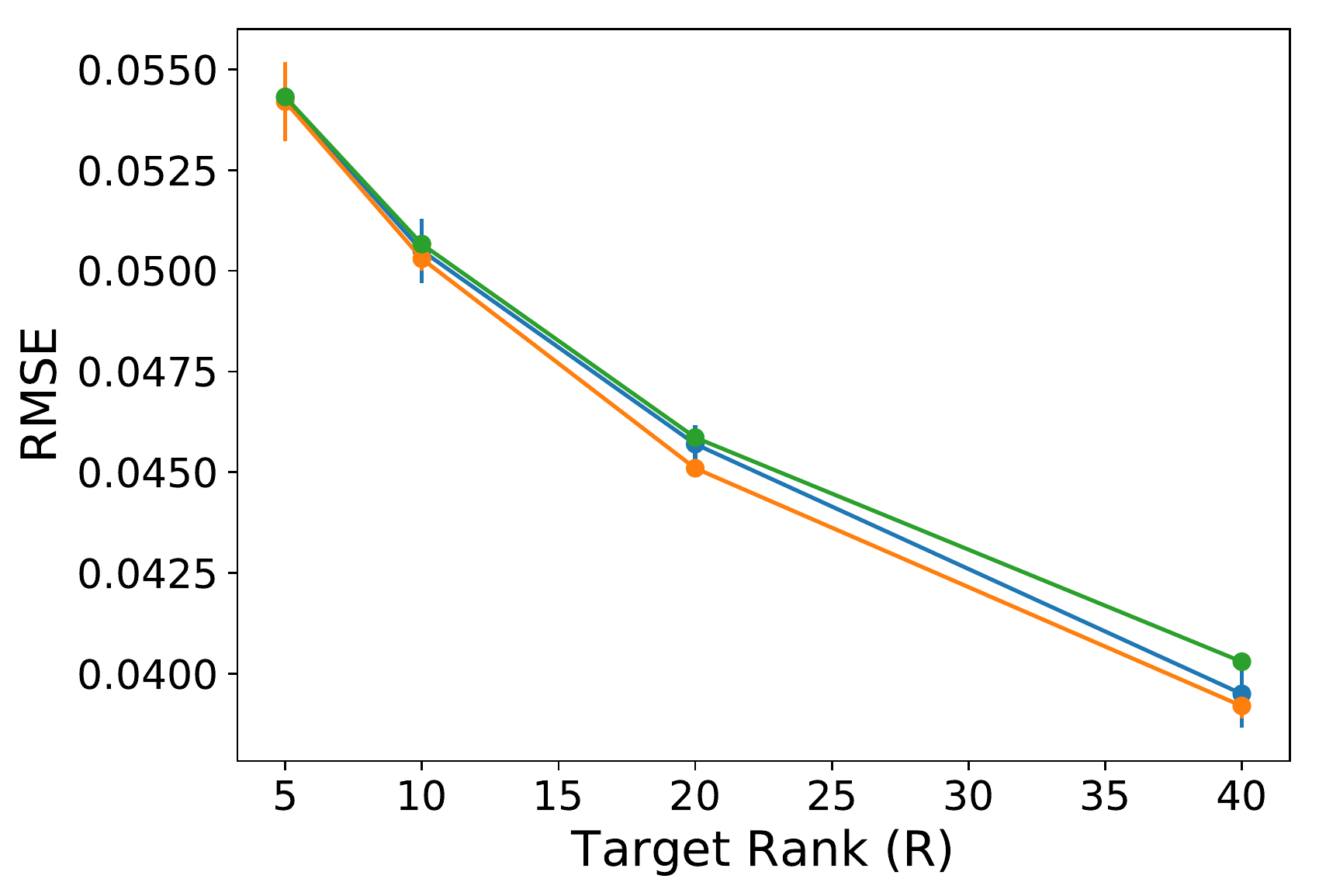}
        \caption{RMSE for Sutter}
        \label{fig:RMSE_Sutter}
    \end{subfigure}%
    \begin{subfigure}[t]{0.37\textwidth}
        \includegraphics[height=1.4 in]{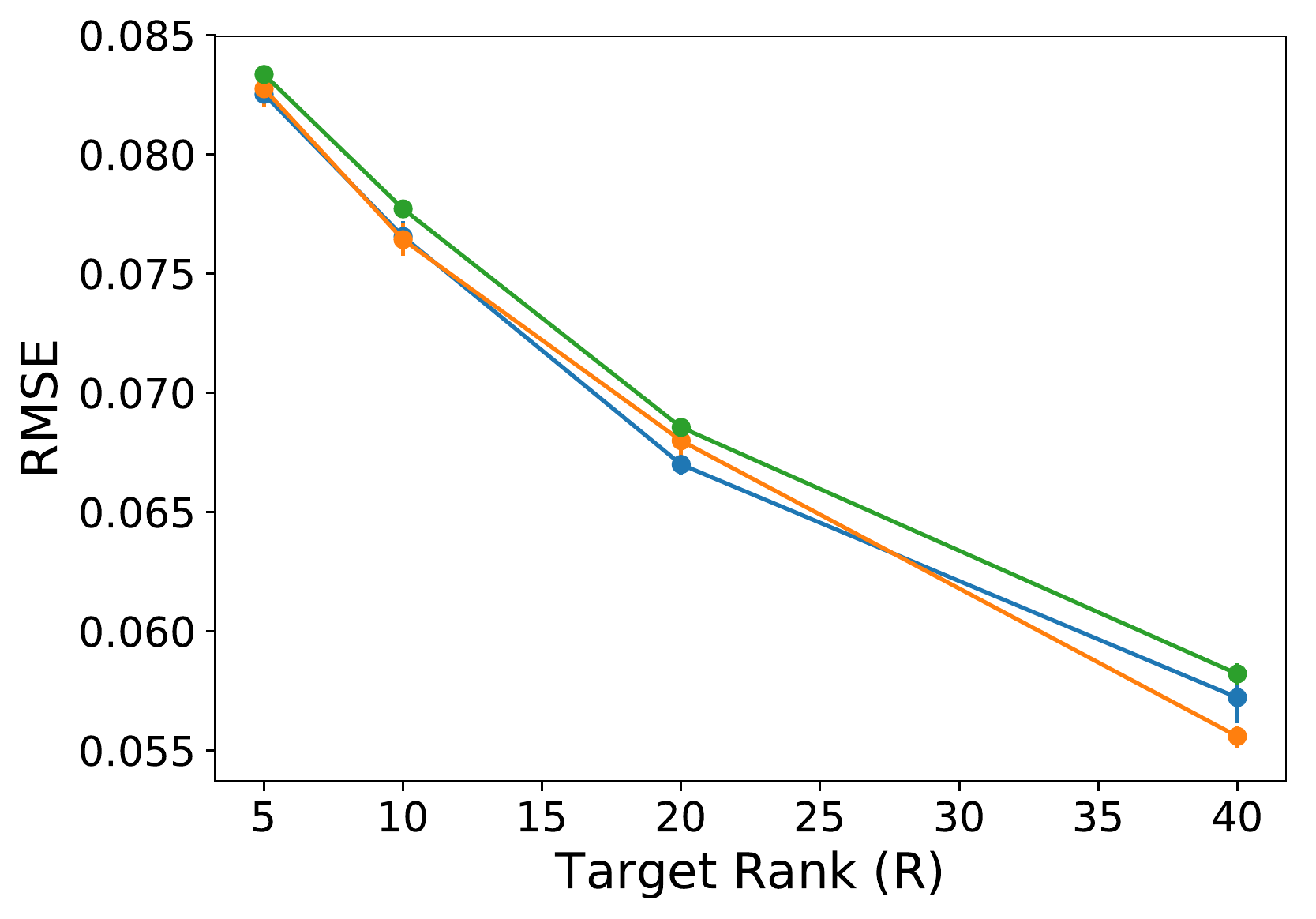}
        \caption{RMSE for CMS}
         \label{fig:RMSE_CMS}
    \end{subfigure}\\ 
    \begin{subfigure}[t]{0.36\textwidth}
        \includegraphics[height=1.4 in]{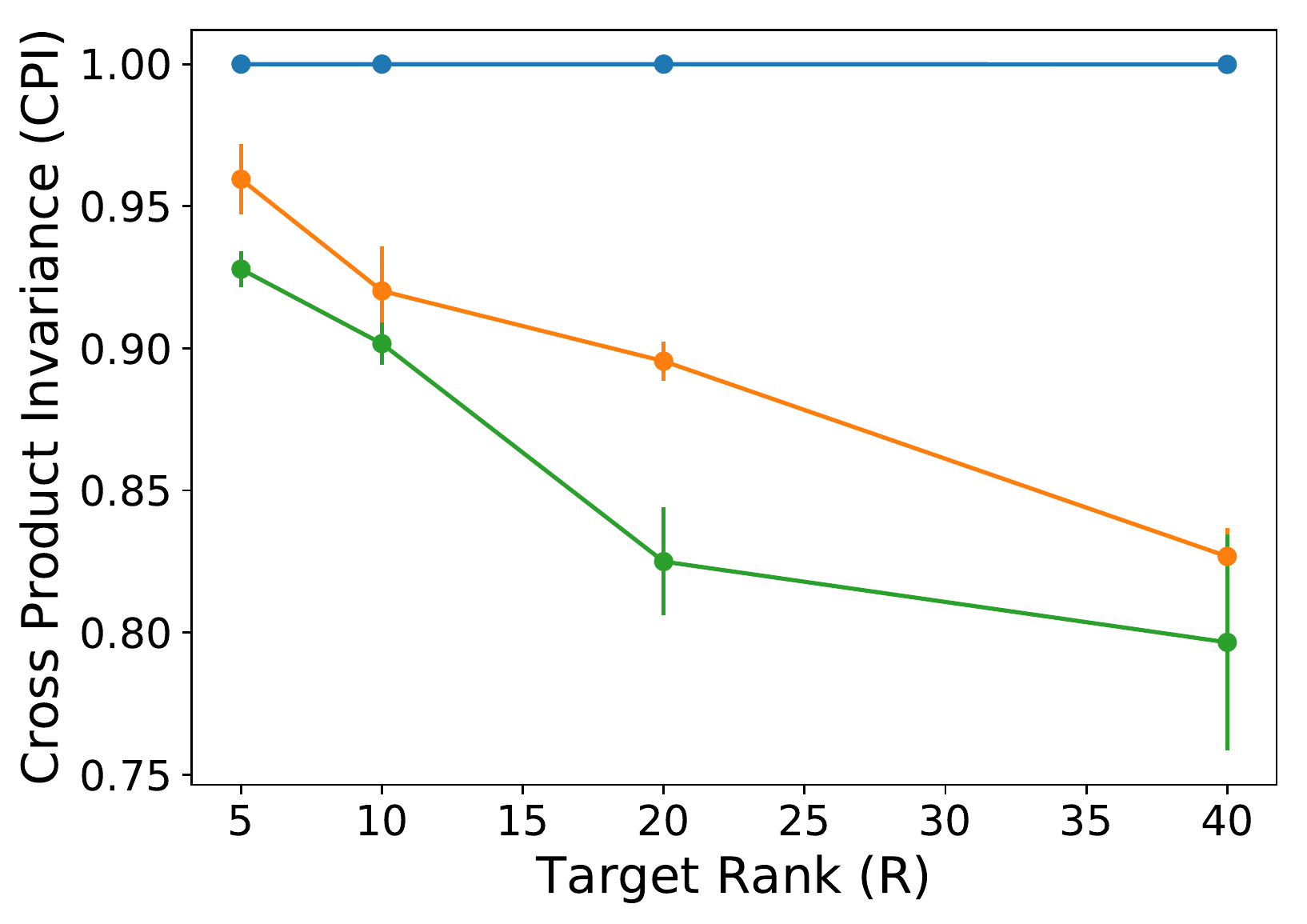}
        \caption{CPI for Sutter}
        \label{fig:CPI_Sutter}
    \end{subfigure}%
    \begin{subfigure}[t]{0.36\textwidth}
        \includegraphics[height=1.4 in]{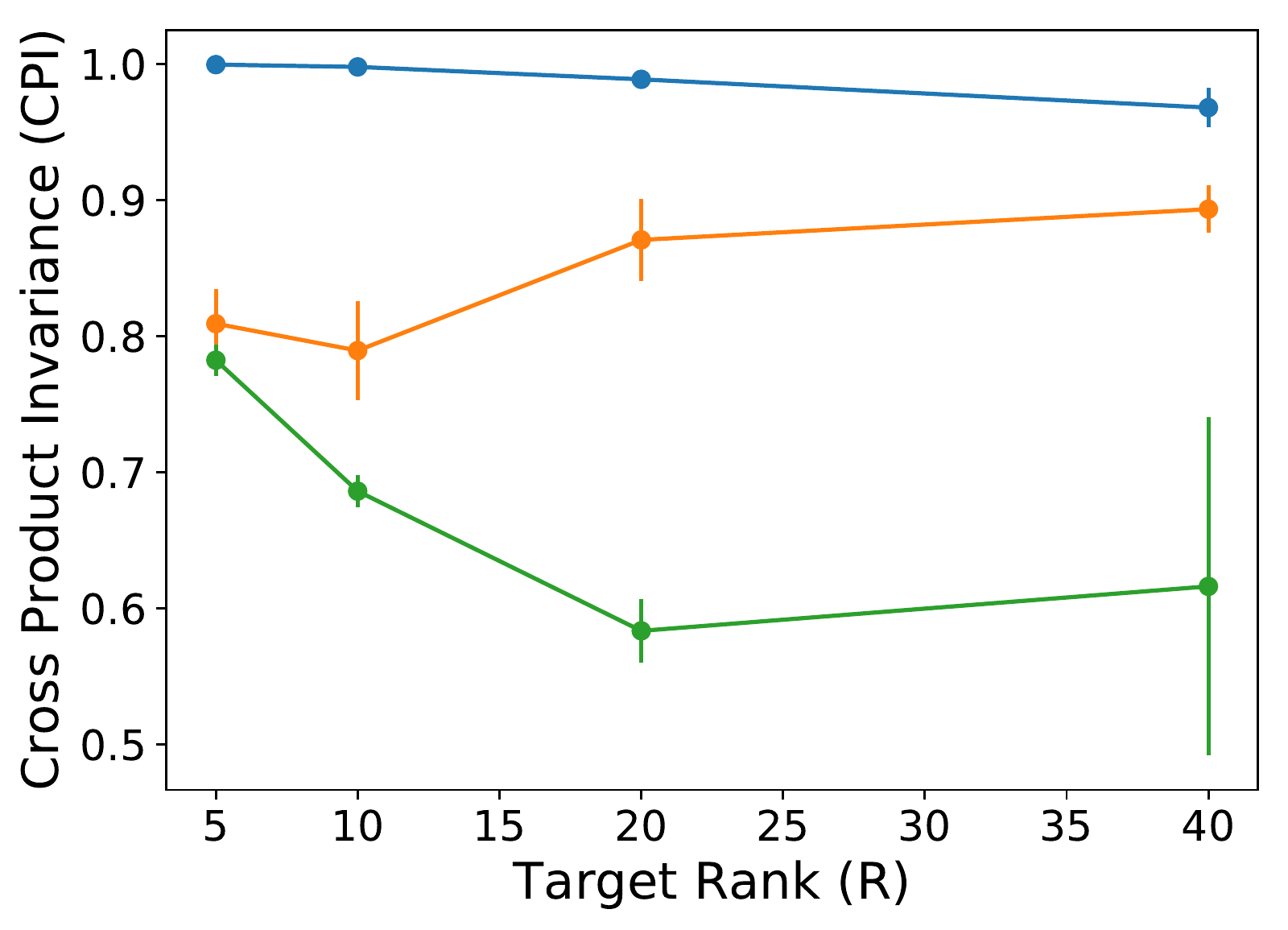}
        \caption{CPI for CMS}
         \label{fig:CPI_CMS}
    \end{subfigure}\\
    \begin{subfigure}[t]{0.37\textwidth}
        \includegraphics[height=1.4 in]{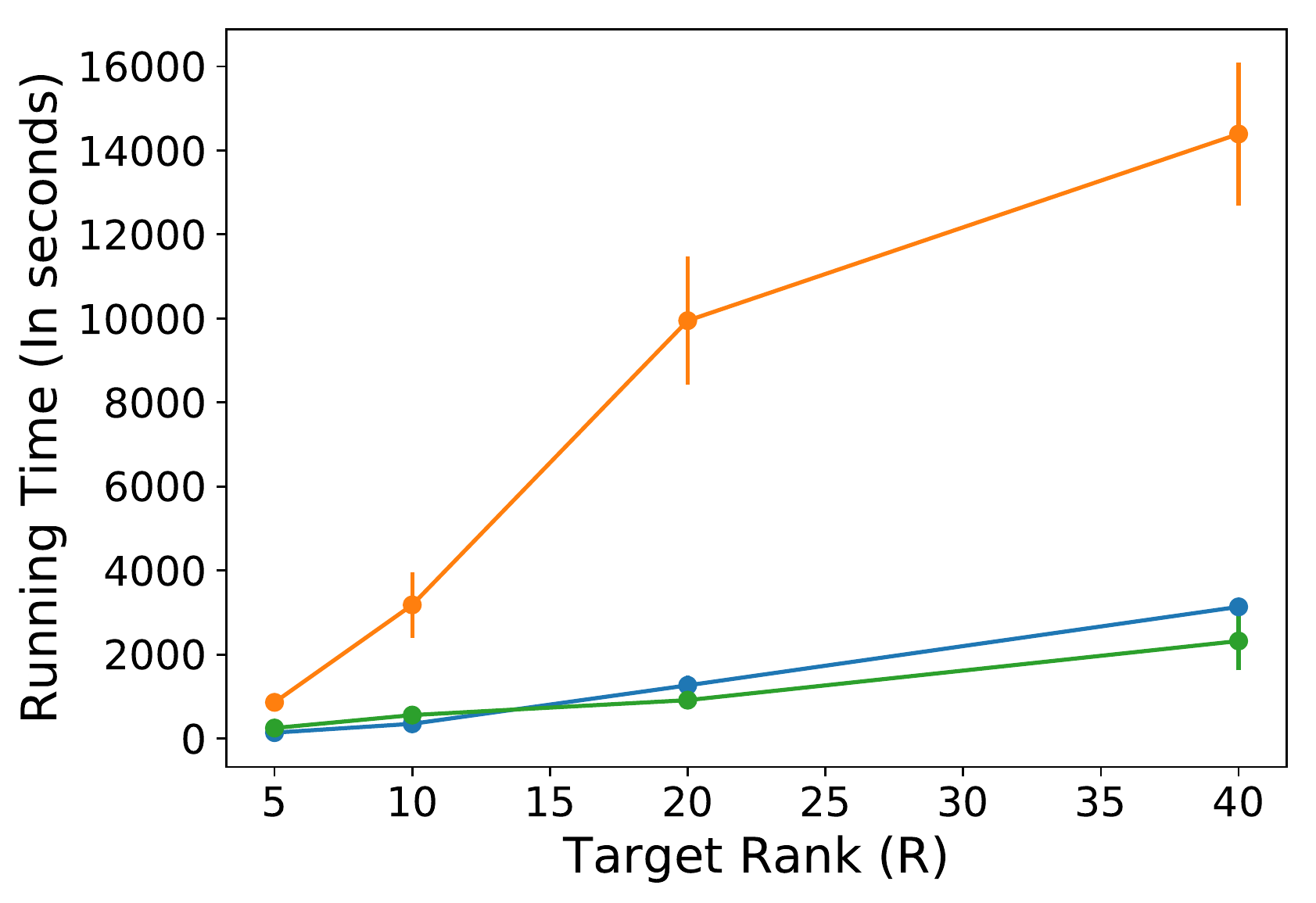}
        \caption{Running Time for Sutter}
        \label{fig:TIME_Sutter}
    \end{subfigure}%
    \begin{subfigure}[t]{0.37\textwidth}
        \includegraphics[height=1.4 in]{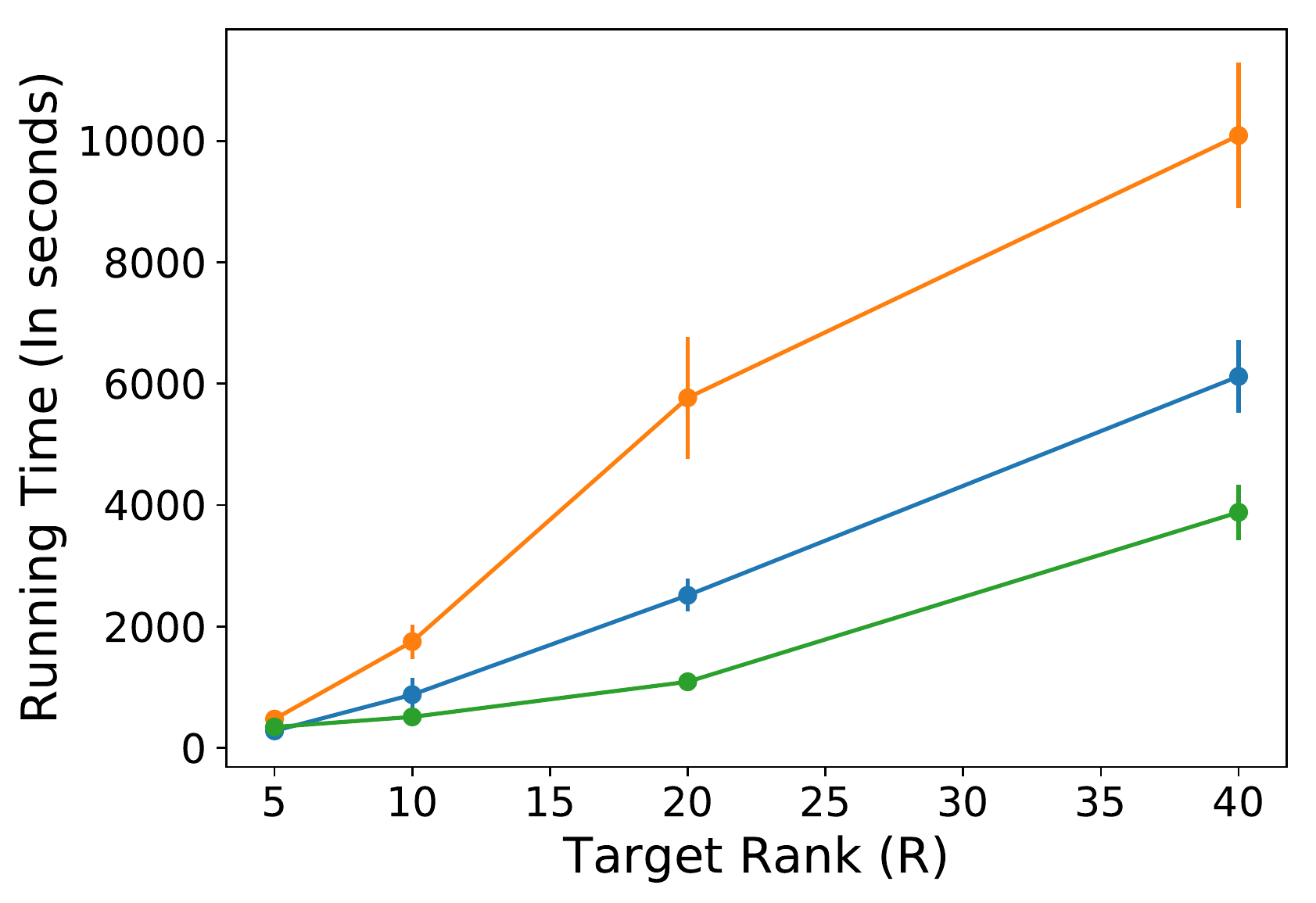}
        \caption{Running Time for CMS}
         \label{fig:TIME_CMS}
    \end{subfigure}\\

    \caption{  The average and standard deviation of RMSE (lower is better), CPI (higher is better), and  total running time (in seconds) (lower is better) for different approaches and for  different target ranks ($R=\{5,10,20,40\}$) related to 5 different random initialization for Sutter and CMS data sets.}
    \label{fig:RMSE_CPI_TIME_COMP}
\end{figure}

Apart from purely evaluating the RMSE and the computational time achieved, we assess to what extent the cross-product invariance constraint is satisfied ~\cite{kiers1999parafac2}.  Therefore, in Figure \ref{fig:RMSE_CPI_TIME_COMP} we present the average and standard deviation of RMSE, CPI, and the computational time 
for the approaches under comparison for Sutter and CMS data sets for four different target ranks ($R \in \{5, 10, 20, 40\}$).  In Figures ~\ref{fig:RMSE_Sutter}, \ref{fig:RMSE_CMS}, we compare the RMSE for all three methods. We remark that all methods achieve  comparable RMSE values for two different data sets. On the other hand, Figures \ref{fig:CPI_Sutter}, \ref{fig:CPI_CMS} show the cross-product invariance (CPI) for Sutter and CMS respectively. COPA+ achieves poor values of CPI for both data sets. This indicates that the output factors violate model constraints and would not satisfy uniqueness properties \cite{kiers1999parafac2}. %\kp{if we have another experiment specifically assessing solution uniqueness, we may need to refer to that also here.}
%Therefore, we remark that enforcing the non-negativity of $\{\M{U}_k\}$ factors in the heuristic approach as in \mname does not preserve an important model property which promotes unique solution. 
Also \mname outperforms Cohen significantly on CPI in Figures  \ref{fig:CPI_Sutter} and \ref{fig:CPI_CMS}. Finally,  Figures \ref{fig:TIME_Sutter}, \ref{fig:TIME_CMS} show the running time comparison for all three methods on where \mname  is up to $4.5 \times$ and   $2 \times$  faster than Cohen on Sutter and CMS data sets. Therefore, our approach is the only one that achieves a fast and accurate solution (in terms of RMSE) and preserves model uniqueness (in terms of CPI).% \kp{why do we have diff numbers here as compared to intro/abstract? are these average gains? what are the max gains? .}

%\kp{keep in mind to update those with RMSE and be precise about speedups based on new results}

%Overall, both variations of our proposed framework achieve faster computation for the same level of accuracy, as compared to the baseline we created based on~\cite{cohen2018nonnegative}. In particular, 
%\methodNameA is up to $5.6~\times$ faster on Sutter and $8.5~\times$ faster on CMS data set. Also, \methodNameB achieves up to $23~\times$ and $18~\times$ speed-up over the modified approach based on~\cite{cohen2018nonnegative} on Sutter and CMS data sets, respectively. %Both of our proposed models are 11 $\%$ more accurate than the baseline.

%\subsection{Cross-product invariance measure}

%\subsection{Uniqueness Property of \mname}
%\js{why is this not answering a specific question we had? we should probably split Q1 into 2 questions so scalable is one separate question.}

\subsection{\textbf{Q2. \mname is scalable }
} \label{sec:exp_scale}
\begin{figure}[t]
    \centering
    \begin{subfigure}[t]{0.35\textwidth}
        \includegraphics[height=0.25 in]{results/label.pdf}
         \label{fig:label}
    \end{subfigure}\\
    \begin{subfigure}[t]{0.5\textwidth}
    \includegraphics[height=1.7 in]{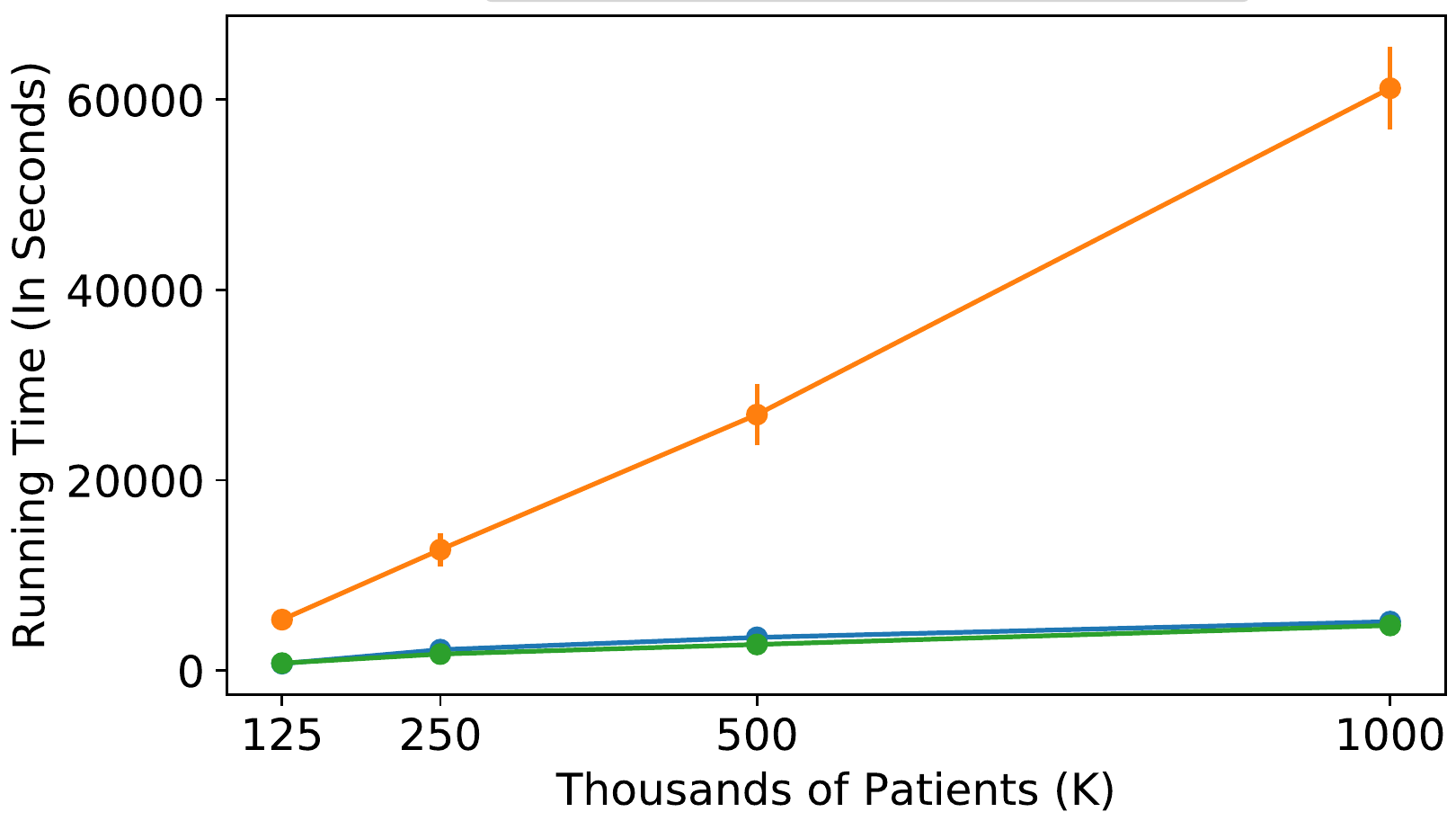} 
    
     \end{subfigure}
     \caption{   The average and standard deviation of running time  (in seconds) for $R=40$ and for 5 random initialization by varying number of patients from 125K to 1 million for CMS data set. \mname is upto $14 \times $  faster than Cohen.
}
\label{fig:scalability_CMS}
\end{figure}

%\kp{i would group RMSE and CPI in one section and Computational time into another (this one). }
%\ho{I would group the runtime comparisons all at once... I would say it's fast and scalable rather than fast, accurate. then another section on scalability? Personal preference though.}
 Apart from assessing the time needed for increasing values of target rank (i.e., number of phenotypes), we evaluate the approaches by comparison in terms of time needed for increasing load of input patients. Each method runs for $5$ times and the convergence threshold is set to $1e-4$ for all of them.   Figure \ref{fig:scalability_CMS} compares the average and standard deviation of total running time for 125K, 250K, 500K, and 1 Million patients for $R=40$. \mname is  up to  $14 \times$ faster than Cohen's baseline for $R=40$. While COPA+ is  a fast approach, this baseline suffers from not satisfying model constraints which promote uniqueness. %as we demonstrated in previous Sections.

%\kp{the paragraph above needs some work, i am doing an attempt to rewrite, Ari please make sure that it makes sense: Apart from assessing the time needed for increasing values of target rank (i.e., number of phenotypes), we evaluate the approaches under comparison in terms of time needed for increasing load of input patients. Each method is run $3$ times and the convergence threshold is set to $1e-4$ for all of them. We provide the results of this experiment in Figure \ref{fig:scalability_comp} for two different values of target rank ($R=\{5, 40\}$). \textit{Adjust the last sentence as needed, pls report both average and max gains.}}

\subsection{Q3. Static features in \mname improve predictive power}
We measure the performance of \mname indirectly on the performance of classification. So we predict whether a patient will be diagnosed with heart failure (HF) or not. Our objective is to assess whether static features in the way handled by \mname boost predictive performance by using personalized phenotype scores for all patients (\M{W}) as features.
\subsubsection{\textbf{Cohort Construction:}}  After applying the preprocessing steps (i.e. removing sparse features and eliminating patients with less than 5 clinical visits), we create a data set with 35113 patients where 3244 of them are cases and 31869 related to controls (9.2 \% are cases).  For case patients, we know the date that they are diagnosed with heart failure (HF dx). Control patients also have the same index dates as their corresponding cases. We extract 145 medications and 178 diagnosis codes from a 2-year observation window and set prediction window length to 6 months. Figure \ref{fig_obs_wind} in the supplementary material section depicts the observation and prediction windows in more detail.

\subsubsection{\textbf{Baselines:}}
We assess the performance of \mname with 6 different baselines. 
\begin{itemize}
    \item \textbf{RNN-regularized CNTF:}
    %\footnote{We contacted the author of the paper, but the code wasn't available so we implemented their method. \kp{I do not think it is useful to state that here.}}
    CNTF \cite{YQCFP19aaai} feeds the temporal phenotype evolution matrices ($\{\M{U_k}\}$) into an LSTM model for HF prediction. This baseline only uses  temporal medical features.
    \item \textbf{RNN Baseline:} We use the GRU model for HF prediction implemented in \cite{choi2016using}. The one-hot vector format is used to represent all dynamic and static features for different clinical visits. 
    \item \textbf{Logistic regression with raw dynamic:}  We  create a binary matrix where the rows are the number of patients  and columns are the total number of medical features (323). Row $k$ of this matrix is created  by aggregating over all clinical visits of matrix \M{X_k}.
    \item \textbf{Logistic regression with raw static+dynamic:} Same as the previous approach, we create a binary matrix where the rows are number of patients and columns are the total number of temporal and static features (345) by appending matrix \M{A} to raw dynamic baseline matrix.

    %\js{rename baseline1 and baseline2 to raw static+dynamic, raw dynamic. make the corresponding change in the figure as well}
        \item \textbf{COPA Personalized Score Matrix:} We use the implementation of pure PARAFAC2 from \cite{afshar2018copa} which learn the low-rank representation of phenotypes (\M{V_{copa}}) from training set and then projecting all the new patients onto the learned phenotypic low-rank space.
        \item \textbf{COPA (+static) Personalized Score Matrix:} This is same as previous baseline, however, we incorporate the static features into PARAFAC2 matrix by repeating the value of static features of a particular patient for all  encounter visits.
\end{itemize}


\begin{figure}[htb]
\centering
\includegraphics[scale=0.65]{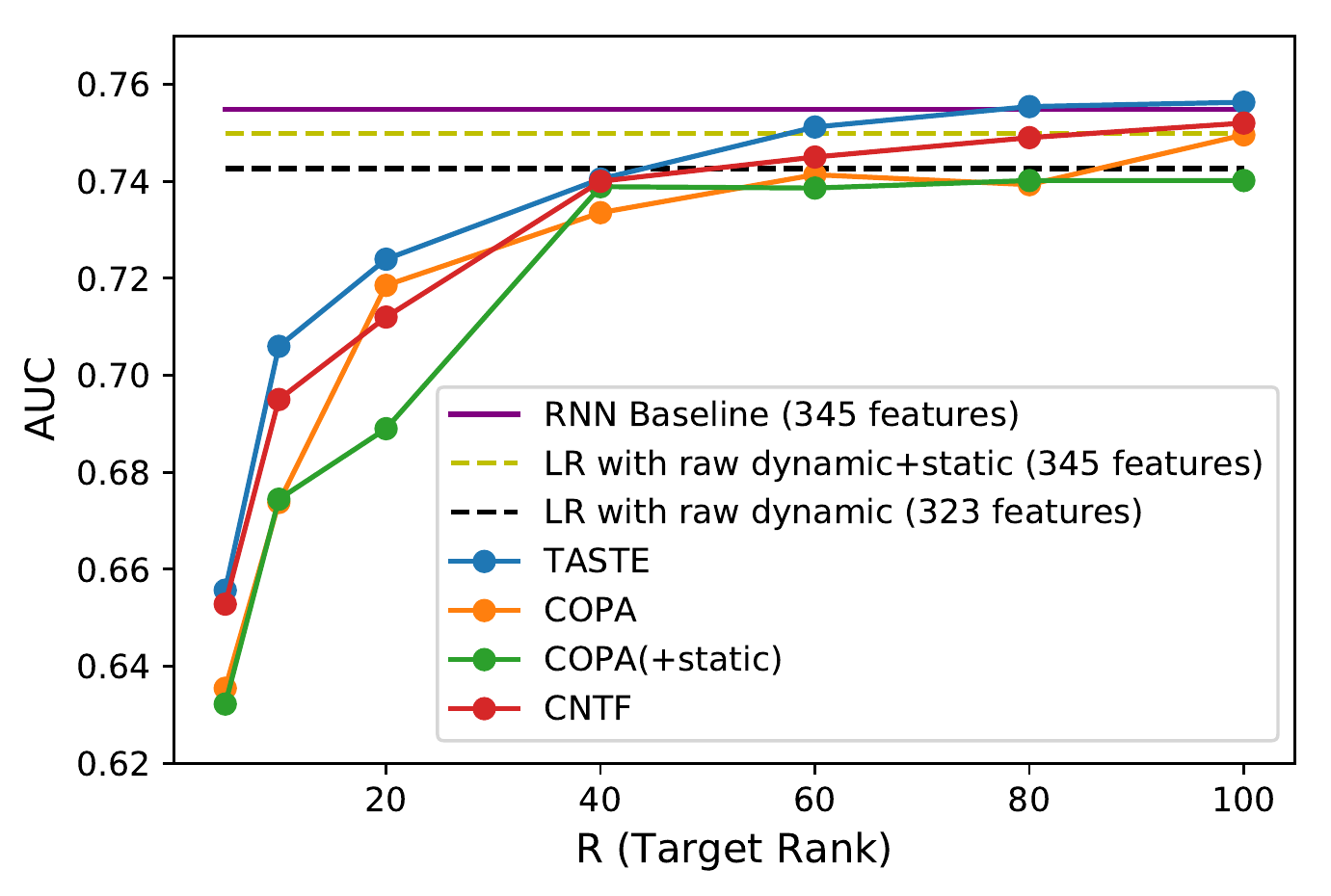}
\caption{  The average of AUC score for varying number of phenotypes (R) for \mname and 3 other tensor baselines on the test set. The AUC score for a baseline with raw dynamic features (323) is \textbf{0.7426}, for the raw dynamic+static baseline (345) is \textbf{0.7498} and for RNN baseline is \textbf{0.7547}. 
}
\label{fig_AUC_score}
\end{figure}

\paragraph{\textbf{Results:}}
Figure~\ref{fig_AUC_score} shows the average of AUC for  all baselines and \mname. For COPA, COPA(+static), CNTF and \mname we report the AUC score for different values of R (\{5,10,20,40,60,80,100\}). \mname improves the AUC score over a simple non-negative PARAFAC2 model (COPA and COPA(+static)) and CNTF which suggests: 1) incorporating static features with dynamic ones will increase the predictive power (comparison of \mname with COPA and CNTF).  2) incorporating static features in the way we do in \mname improves predictive power (comparison of \mname and COPA(+static)). %The average  AUC score for the RNN baseline is $0.7547$ and for the lasso logistic regression baseline with raw dynamic features (323) is $0.7426$ and for a baseline with raw dynamic and static features (345) is $0.7498$. %\ho{If you need space, remove the average AUC since it's in the figure caption. Also the numbers are not consistent with caption (the number of features)}
\mname with R=80 (AUC=$0.7554$) performs slightly better than the RNN model. Training details are provided in the supplementary material section. %For R=40, 80,  \mname is able to get  comparable AUC as raw dynamic and raw static+dynamic baselines. 

%\kp{let's keep in mind to re-run case study as well, based on a more thorough hyper-param search choosing based on rmse. hopefully the result in table 3 will be reproducible.} 
\subsection{Q4. \textbf{Heart Failure Phenotype Discovery}} \label{case_study}

%\kp{I think we need to motivate the use of PARAFAC2 for this task instead of e.g., collapsing the time mode through aggregation. Are there any examples of patient temporal signatures worth highlighting? We probably need to extract some examples and try to get Chris' feedback on the clinical meaningfulness}
Heart failure (HF) is a complex, heterogeneous disease and is the leading cause of hospitalization in people older than 65~\footnote{\url{https://www.webmd.com/heart-disease/guide/diseases-cardiovascular\#1-4}}. However, there are no well-defined phenotypes other than the simple categorization of ejection fraction of the heart (i.e., preserved or reduced ejection fraction). % \kp{can we find a citation to support that?}.
With the comprehensive collection of available longitudinal EHR data, now we have the opportunity to computationally tackle the challenge of phenotyping HF patients. 

 \subsubsection{\textbf{Cohort Construction:}} 
 We select the patients diagnosed with HF from the EHRs in Sutter dataset. We extract 145 medications and 178 diagnosis codes from a 2-year observation window which ends 6 months before the heart failure diagnosis date (HFdx)\footnote{Figure \ref{fig_obs_wind} in the supplementary material section presents the observation  window in more detail.}. The total number of patients (K) is 3,244 (the HF case patients of Sutter dataset).

\subsubsection{\textbf{Findings of HF Phenotypes}}
\mname extracted 5 phenotypes which are all confirmed and annotated by an expert cardiologist \footnote{More detail about the strategy for finding the optimal number of phenotypes is provided in the supplementary material section.}. The clinical description of all phenotype are provided by the cardiologist:
\begin{enumerate}
    \item [P1.] \textbf{Atrial Fibrillation (AF)}: This phenotype represents patients with irregular heartbeat and AF predisposes to HF. Medications are related to managing AF and preventing strokes. This phenotype is usually more prevalent in male and old patients (i.e. 80 years or older). 
    \item [P2.] \textbf{Hypertensive Heart Failure:}
    This is a classic and dominant heart failure phenotype, representing a subgroup of patients with long history of hypertension, and cardiac performance declines over time. Anti-hypertensive medications are spelled out as to indicate the treatment to hypertension.
    \item [P3.] \textbf{Obese Induced Heart Failure:} This phenotype is featured by severe obesity (BMI>35) and obesity induced orthopedic  condition. %The phenotype is consistent with clinical phenotypes though the underlying pathophysiology is not quite clear.
    \item [P4.] \textbf{Cardiometablic Driving Heart Failure:} This phenotype is featured by diabetes and cardiometabolic conditions (i.e. hyperlipidemia, hypertension). Diabetes is a well known risk factor for cardiovascular complications (i.e. stroke, myocardial infaction, etc.), and increases the risk for heart failure.   
    \item [P5.] \textbf{Coronary Heart Disease Phenotype:} This phenotype is associated with a greater deterioration of left ventricle function and a worse prognosis. This phenotype is also more prevalent in male and white population.  
\end{enumerate}
More detail about all phenotypes is provided in the supplementary material section.

%% file: 05-Conclusion.tex
\section{Conclusions}
%Through this work, we 
\mname~jointly models temporal and static information from electronic health records to extract clinically meaningful phenotypes.
We demonstrate the computational efficiency of our model on extensive experiments that showcase its ability to preserve important properties underpinning the model's uniqueness, while maintaining interpretability. 
%We introduce a fast and accurate algorithm to fit our model and demonstrate its computational efficiency and its ability to preserve important properties underpinning the model's uniqueness, while maintaining interpretability.
%We show incorporating static feature in a way we do in \mname can improve the predictive power. 
\mname not only identifies clinically meaningful heart failure phenotypes validated by a cardiologist but the phenotypes also retain predictive power for predicting heart failure.
%We applied \mname method to identify clinically meaningful heart failure phenotypes validated by a cardiologist. We also showed the high power of phenotypes for predicting heart failure. 

%In this paper, we proposed \methodName which jointly analyze both dynamic and static features for the  number of patients. The dynamic information may have varying number of clinical visits among the patients which can be modeled through PARAFC2.  The static information also can modeled by a matrix. Therefore, \methodName tries to decomposes A PARAFAC2  tensor and a matrix where a factor matrix is shared between them. All the factor matrices including the variable mode ($\{\M{U_k}\}$) is non-negative. The non-negativity constraint on $\{\M{U_k}\}$ may violate the  uniqueness. Therefore, we utilized alternating optimization and alternating direction method of multipliers (AO-ADMM) to carefully make the factor matirces non-negative without violating the uniqueness.   \mname is up to 8.5 $\times$ faster than the baseline method on the real data sets. 

%% file: 06-Acknowledgements.tex
\section{Acknowledgements}
This work was in part supported by the National Science Foundation award IIS-1418511, CCF-1533768 and IIS-1838042, the National Institute of Health award 1R01MD011682-01 and R56HL138415.

%% file: 06-suplemntary_matrials.tex
\clearpage
\section{Supplementary Material}

\subsection{Non-Negativity constrained Least Squares (NNLS)} NNLS problem  has the following form:

\begin{equation}
\small
\begin{aligned}
& \underset{\M{C}}{\text{minimize}}
& & \norm{\M{B}\M{C^T}-\M{A}}_F^2 \\
& \text{subject to}
& &  \M{C} \ge 0
\end{aligned}
\label{NNLS_sub}
\end{equation}

Here, $\M{A} \in \mathbb{R}^{M \times N}$, $\M{B} \in \mathbb{R}^{M \times R}$ and $\M{C} \in \mathbb{R}^{N \times R}$ where $R \ll min(M, N)$. NNLS is a convex problem and the optimal solution of \ref{NNLS_sub} can be easily found. In this paper, we use the block principal pivoting method \cite{kim2011fast} to solve NNLS problems. Authors showed block principal pivoting method has the state-of-the-art performance.

\subsection{More details on \mname framework}
\subsubsection{Updating factor matrix \M{Q_k}}

We can rewrite  objective function \ref{NN_coupled_PARAFAC2}  with respect to $\M{Q_k}$ based on trace properties~\cite{petersen2008matrix} as:  

\begin{equation}
\begin{aligned}
& \underset{\M{Q_k}}{\text{minimize}}
& &  \underbrace{\frac{\mu_k}{2}\text{Trace}(\M{U_k}^T\M{U_k})}_\text{constant} -\mu_k \text{Trace}(\M{U_k}^T\M{Q_k}\M{H})\\
& & & +\underbrace{\frac{\mu_k}{2} \text{Trace}(\M{H}^T\M{Q_k}^T\M{Q_k}\M{H})}_\text{constant}\\
& \text{subject to}
& &  \M{Q_k^T} \M{Q_k}=\M{I}   
\end{aligned}
\label{NN_coupled_PARAFAC2_Lagrangian}
\end{equation}

After removing the constant terms and applying $\text{Trace}(\M{ABC})=\text{Trace}(\M{CAB})$, we have:

\begin{equation}
\small
\begin{aligned}
& \underset{\M{Q_k}}{\text{minimize}}
& & \mu_k\norm{  \M{U_k} \M{H}^T-\M{Q_k}}_F^2 \\
& \text{subject to}
& &  \M{Q_k^T} \M{Q_k}=\M{I}
\end{aligned}
%\label{Equation_Q_k}
\end{equation}

\subsubsection{Updating factor matrix \M{H}} \label{update_H_details}
$\M{Q_k} \in \mathbb{R}^{I_k \times R}$ is a rectangular orthogonal matrix ($\M{Q_k}^T \M{Q_k}=\M{I} \in \mathbb{R}^{R \times R}$). We also introduce $\widetilde{\M{Q_k}} \in \mathbb{R}^{I_k \times I_k-R}$ where  $\widetilde{\M{Q_k}}^T \widetilde{\M{Q_k}}=\M{I} \in \mathbb{R}^{I_k-R \times I_k-R}$ and $\widetilde{\M{Q_k}}^T \M{Q_k}=\M{0}$. Now $[\M{Q_k} \quad  \widetilde{\M{Q_k}}]$ is a square orthogonal  matrix as follows: 
\begin{equation}
\begin{aligned}
\begin{bmatrix}
\M{Q_k}^T \\
\widetilde{\M{Q_k}}^T
\end{bmatrix} \begin{bmatrix}
\M{Q_k}
\quad
\widetilde{\M{Q_k}}
\end{bmatrix} \\
& =\begin{bmatrix*}[l]
\M{Q_k}^T \M{Q_k}
&
\M{Q_k}^T \widetilde{\M{Q_k}} \\
\widetilde{\M{Q_k}}^T \M{Q_k}
&
 \widetilde{\M{Q_k}}^T \widetilde{\M{Q_k}}
\end{bmatrix*}\\
& =\begin{bmatrix*}[l]
\M{I}_{R \times R}
&
0 \\
0
&
\M{I}_{(I_k-R) \times (I_k-R)}
\end{bmatrix*}=\M{I}_{I_k \times I_k}
\end{aligned}
\end{equation}

Now we multiply $[\M{Q_k} \quad  \widetilde{\M{Q_k}}]$ with equation~\ref{H_equation} as follow:

\begin{equation}
\small
\begin{aligned}
\sum_{k=1}^{K} \frac{\mu_k}{2}\norm{\M{Q_k}\M{H}-\M{U_k}}_F^2\\
& =\sum_{k=1}^{K} \frac{\mu_k}{2}\norm{[\M{Q_k} \quad \widetilde{\M{Q_k}}]^T \Big( \M{Q_k}\M{H}-  \M{U_k}\Big)}_F^2 \\ & =\sum_{k=1}^{K} \frac{\mu_k}{2}\norm{\begin{bmatrix}
\M{Q_k}^T  \\
\widetilde{\M{Q_k}}^T 
\end{bmatrix} \Big(\M{Q_k}\M{H}-\M{U_k} \Big) }_F^2\\
&=\sum_{k=1}^{K} \frac{\mu_k}{2}\norm{\begin{bmatrix}
\M{Q_k}^T \M{Q_k}  \\
\widetilde{\M{Q_k}}^T \M{Q_k} 
\end{bmatrix} \M{H} -\begin{bmatrix}
\M{Q_k}^T \M{U_k} \\
\widetilde{\M{Q_k}}^T \M{U_k}
\end{bmatrix}}_F^2 \\
& =\sum_{k=1}^{K} \Big( \frac{\mu_k}{2}\norm{\M{H}-\M{Q_k}^T \M{U_k}}_F^2+\overbrace{\norm{\widetilde{\M{Q_k}}^T \M{U_k}}_F^2}^\text{constant}\Big) \\
%& =\sum_{k=1}^{K} \frac{\mu_k}{2}\norm{\M{H}-\M{Q_k}^T \M{U_k}}_F^2
\label{H_equation}
\end{aligned}
\end{equation}

where $\sum_{k=1}^{K} \norm{\widetilde{\M{Q_k}}^T \M{U_k}}_F^2$ is a constant and independent of the parameter under minimization. Therefore, the value of \M{H} that minimizes $\sum_{k=1}^{K} \frac{\mu_k}{2}\norm{\M{Q_k}\M{H}-\M{U_k}}_F^2$ also minimizes $\sum_{k=1}^{K}  \frac{\mu_k}{2}\norm{\M{H}-\M{Q_k}^T \M{U_k}}_F^2$

%\subsection{\textbf{Setup}}
%\subsubsection{\textbf{Data Set Description}}\label{data_describ}

\subsection{\textbf{Implementation details}}
\mname is implemented in MATLAB. To enable reproducibility of our work, we attached a zip file containing \mname code. 
All the approaches (including the baselines) are evaluated on MatlabR2017b. We utilize the capabilities of Parallel Computing Toolbox of Matlab by activating parallel pool for all methods. For Sutter and CMS  dataset, we used 12 workers. For the prediction task, we use the implementation of regularized logistic regression from Scikit-learn machine learning library of python 3.6.

\begin{table}[htbp]
  \centering
  \caption{\footnotesize \mname extracted 5 phenotypes from the HF dataset. \textcolor[rgb]{ .867,  .031,  .024}{Red} indicates the static features; $Dx\_$ indicates diagnoses;  $Rx\_$ indicates medication; \textbf{The phenotype names are provided by the cardiologist.}}
  \scalebox{0.75}{
    \begin{tabular}{lr}
    \textbf{ P1. Atrial Fibrillation (AF)} & \multicolumn{1}{l}{\textbf{Weight}} \\
    \midrule
    Dx\_Cardiac dysrhythmias [106.] & 1 \\
    Rx\_Coumarin Anticoagulants & 0.79597 \\
    Dx\_Heart valve disorders [96.] & 0.57526 \\
    Dx\_Coronary atherosclerosis and other heart disease [101.] & 0.465406 \\
    Rx\_Beta Blockers Cardio-Selective & 0.348005 \\
    Dx\_Conduction disorders [105.] & 0.246467 \\
    \textcolor[rgb]{ .867,  .031,  .024}{Static\_white} & \textcolor[rgb]{ .867,  .031,  .024}{0.270443} \\
    \textcolor[rgb]{ .867,  .031,  .024}{Static\_age\_greater\_80} & \textcolor[rgb]{ .867,  .031,  .024}{0.245424} \\
    \textcolor[rgb]{ .867,  .031,  .024}{Static\_Non\_Hispanic} & \textcolor[rgb]{ .867,  .031,  .024}{0.245373} \\
    \textcolor[rgb]{ .867,  .031,  .024}{Static\_male} & \textcolor[rgb]{ .867,  .031,  .024}{0.200464} \\
    \textcolor[rgb]{ .867,  .031,  .024}{Static\_Alchohol\_yes} & \textcolor[rgb]{ .867,  .031,  .024}{0.197431} \\
    \textcolor[rgb]{ .867,  .031,  .024}{Static\_Smk\_Quit} & \textcolor[rgb]{ .867,  .031,  .024}{0.167838} \\
          &  \\
    \textbf{P2. Hypertensive Heart Failure:} & \multicolumn{1}{l}{\textbf{Weight}} \\
    \midrule
    Dx\_Essential hypertension [98.] & 0.705933 \\
    Rx\_Calcium Channel Blockers & 0.684211 \\
    Rx\_ACE Inhibitors & 0.578426 \\
    Rx\_Beta Blockers Cardio-Selective & 0.564413 \\
    Rx\_Angiotensin II Receptor Antagonists & 0.250883 \\
    Dx\_Chronic kidney disease [158.] & 0.14528 \\
    Rx\_Thiazides and Thiazide-Like Diuretics & 0.134521 \\
    \textcolor[rgb]{ .867,  .031,  .024}{Static\_Non\_Hispanic} & \textcolor[rgb]{ .867,  .031,  .024}{0.536399} \\
    \textcolor[rgb]{ .867,  .031,  .024}{Static\_female} & \textcolor[rgb]{ .867,  .031,  .024}{0.390343} \\
    \textcolor[rgb]{ .867,  .031,  .024}{Static\_Smk\_NO} & \textcolor[rgb]{ .867,  .031,  .024}{0.363068} \\
    \textcolor[rgb]{ .867,  .031,  .024}{Static\_Alchohol\_No} & \textcolor[rgb]{ .867,  .031,  .024}{0.360158} \\
    \textcolor[rgb]{ .867,  .031,  .024}{Static\_white} & \textcolor[rgb]{ .867,  .031,  .024}{0.345942} \\
    \textcolor[rgb]{ .867,  .031,  .024}{Static\_Alchohol\_yes} & \textcolor[rgb]{ .867,  .031,  .024}{0.264664} \\
    \textcolor[rgb]{ .867,  .031,  .024}{Static\_age\_between\_70\_79} & \textcolor[rgb]{ .867,  .031,  .024}{0.26237} \\
    \textcolor[rgb]{ .867,  .031,  .024}{Static\_Severely\_obese} & \textcolor[rgb]{ .867,  .031,  .024}{0.252981} \\
    \textbf{P3. Obese Induced Heart Failure:} & \multicolumn{1}{l}{\textbf{Weight}} \\
    \midrule
    Dx\_Other back problems & 0.438879 \\
    Rx\_Opioid Combinations & 0.414518 \\
    Dx\_Other connective tissue disease [211.] & 0.304155 \\
    Dx\_Other non-traumatic joint disorders [204.] & 0.269159 \\
    Dx\_Osteoarthritis [203.] & 0.190075 \\
    Rx\_Opioid Agonists & 0.185262 \\
    Dx\_Intervertebral disc disorders & 0.156421 \\
    Rx\_Benzodiazepines & 0.150865 \\
    \textcolor[rgb]{ .867,  .031,  .024}{Static\_female} & \textcolor[rgb]{ .867,  .031,  .024}{0.548758} \\
    \textcolor[rgb]{ .867,  .031,  .024}{Static\_Non\_Hispanic} & \textcolor[rgb]{ .867,  .031,  .024}{0.52183} \\
    \textcolor[rgb]{ .867,  .031,  .024}{Static\_white} & \textcolor[rgb]{ .867,  .031,  .024}{0.474545} \\
    \textcolor[rgb]{ .867,  .031,  .024}{Static\_Alchohol\_No} & \textcolor[rgb]{ .867,  .031,  .024}{0.404191} \\
    \textcolor[rgb]{ .867,  .031,  .024}{Static\_Smk\_Quit} & \textcolor[rgb]{ .867,  .031,  .024}{0.273576} \\
    \textcolor[rgb]{ .867,  .031,  .024}{Static\_Severely\_obese} & \textcolor[rgb]{ .867,  .031,  .024}{0.226485} \\
    \textcolor[rgb]{ .867,  .031,  .024}{Static\_age\_between\_70\_79} & \textcolor[rgb]{ .867,  .031,  .024}{0.180683} \\
    \textcolor[rgb]{ .867,  .031,  .024}{Static\_age\_between\_60\_69} & \textcolor[rgb]{ .867,  .031,  .024}{0.180006} \\
          &  \\
    \textbf{P4. Cardiometablic Driving Heart Failure:} & \multicolumn{1}{l}{\textbf{Weight}} \\
    \midrule
    Dx\_Diabetes mellitus without complication [49.] & 0.712628 \\
    Dx\_Disorders of lipid metabolism [53.] & 0.507946 \\
    Rx\_Biguanides & 0.48821 \\
    Dx\_Immunizations and screening for infectious disease [10.] & 0.358349 \\
    Rx\_Diagnostic Tests & 0.353441 \\
    \textcolor[rgb]{ .867,  .031,  .024}{Static\_Alchohol\_No} & \textcolor[rgb]{ .867,  .031,  .024}{0.247165} \\
    \textcolor[rgb]{ .867,  .031,  .024}{Static\_Non\_Hispanic} & \textcolor[rgb]{ .867,  .031,  .024}{0.190547} \\
    \textcolor[rgb]{ .867,  .031,  .024}{Static\_Smk\_NO} & \textcolor[rgb]{ .867,  .031,  .024}{0.184782} \\
    \textcolor[rgb]{ .867,  .031,  .024}{Static\_male} & \textcolor[rgb]{ .867,  .031,  .024}{0.164795} \\
    \textcolor[rgb]{ .867,  .031,  .024}{Static\_white} & \textcolor[rgb]{ .867,  .031,  .024}{0.15693} \\
    \textcolor[rgb]{ .867,  .031,  .024}{Static\_age\_between\_60\_69} & \textcolor[rgb]{ .867,  .031,  .024}{0.131118} \\
          &  \\
          &  \\
    \textbf{P5. Coronary Heart Disease Phenotype:} & \multicolumn{1}{l}{\textbf{Weight}} \\
    \midrule
    Dx\_Disorders of lipid metabolism [53.] & 0.984987 \\
    Rx\_HMG CoA Reductase Inhibitors & 0.442637 \\
    Dx\_Coronary atherosclerosis and other heart disease [101.] & 0.265414 \\
    Dx\_Diabetes with renal manifestations & 0.229549 \\
    Dx\_Chronic kidney disease [158.] & 0.203386 \\
    Dx\_Other thyroid disorders & 0.101615 \\
    \textcolor[rgb]{ .867,  .031,  .024}{Static\_Non\_Hispanic} & \textcolor[rgb]{ .867,  .031,  .024}{1} \\
    \textcolor[rgb]{ .867,  .031,  .024}{Static\_white} & \textcolor[rgb]{ .867,  .031,  .024}{0.998604} \\
    \textcolor[rgb]{ .867,  .031,  .024}{Static\_male} & \textcolor[rgb]{ .867,  .031,  .024}{0.957132} \\
    \textcolor[rgb]{ .867,  .031,  .024}{Static\_Smk\_Quit} & \textcolor[rgb]{ .867,  .031,  .024}{0.773904} \\
    \textcolor[rgb]{ .867,  .031,  .024}{Static\_Alchohol\_yes} & \textcolor[rgb]{ .867,  .031,  .024}{0.604641} \\

    \end{tabular}%
    }
  \label{tab:other_phenotypes}%
\end{table}%

\subsection{\textbf{More details on Q1. \mname is fast, accurate and preserves uniqueness-promoting constraints}}

\subsubsection{\textbf{Setting hyper-parameters:}}
%We did a parameter search for $\lambda$ and $\mu_k$ and we choose the values that produce the best trade-off between fit and cross-product invariance. 
We perform a grid search for $\lambda \in \{0.01, 0.1, 1\}$ and $\mu_1= \dots = \mu_K = \mu  \in \{0.01, 0.1, 1\}$ for  \mname and Cohen+  for different target ranks ($R=\{5, 10, 20, 40\}$) and we run each method with specific parameter for 5 random initialization and pick the best values of $\lambda$ and $\mu$ based on the lowest average RMSE value. For COPA+, we just search for the best value of $\lambda \in \{0.1, 1,  10\}$ since it does not require $\mu$ parameter. % \kp{how come the CoNP2-OST has a $\mu$ value to tune? in the algorithm, it seems that this version corresponds to the case when $\mu=0$}

\subsection{More details on Q3. Static features in \mname improve predictive power}

\subsubsection{\textbf{Training Details:}} We train a Lasso Logistic Regression for all the baselines under comparison. We divide the patient records into training ($75 \%$), validation ($10 \%$), and test sets ($15 \%$) and use them to evaluate all baselines.  Lasso Logistic Regression has regularization parameter ($C=[1e-2,1e-1,1,10, 100, 1000, 10000]$). For all 6 baselines, we just need to tune parameter $C$.  However, for \mname we  need to perform a 3-D grid search over  $\lambda \in \{0.01, 0.1,  1\}$ and $\mu_1 = \mu_2 = \dots = \mu_K = \mu = \in \{0.01, 0.1,  1\}$ and $C$. We first train \mname on training set and learn the phenotype low-rank representation (\M{V}, \M{F}) and then  train Lasso Logistic Regression classifier based on patient low-rank representation ($\M{W_{train}}$) and corresponding labels. Then we project the patients in validation set on phenotype low-rank representation and  learn their personalized phenotype scores $\M{W_{val}}$. Next, we feed %\kp{is feed the right term here? you probably want to describe that the feature vector learnt is used to compute the probability of whether a certain patient will exhibit HF or not, right?}
$\M{W_{val}}$ into a Lasso Logistic Regression and calculate the AUC score. Finally, we pick the best parameters based on the highest AUC score on validation set and report the AUC on test data.
The training phase for COPA, COPA(+static) and CNTF is same as \mname.

\subsection{More details on Q4. \textbf{Heart Failure Phenotype Discovery}}

% Table generated by Excel2LaTeX from sheet 'temporal_phenotypes_R_5'
\begin{figure}[htb]
\centering
\includegraphics[scale=0.4]{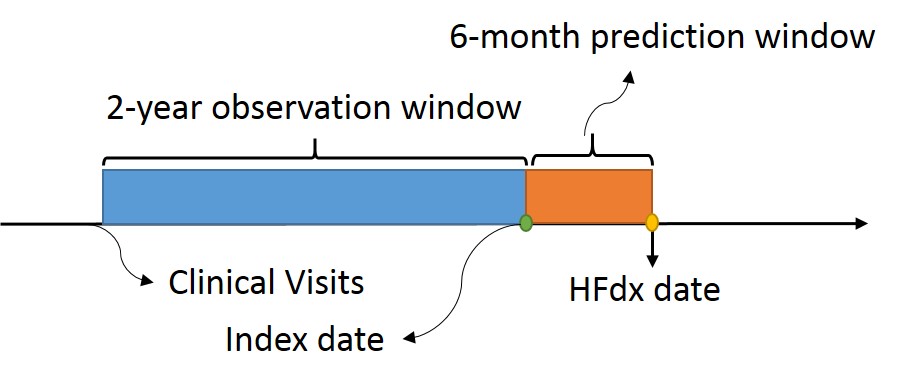}
\caption{ 
The arrow represents the encounter visits of a patient. We extract diagnosis and medications from a 2-year observation window by setting prediction window length to 6 months.
}
\label{fig_obs_wind}
\end{figure}

%\subsubsection{\textbf{Case Study Setup}}
%\ho{Can you spend 1-2 sentences highlighting why HF patients is a good study? It would be good to provide some context.}
%We select the patients diagnosed with HF from the EHRs in D1 dataset. We use an observation window of 12 month  before and after the HF diagnosis date to extract 791 temporal features (medication, diagnosis, procedure) along with their 11 static features as endorsed by a cardiologist.  The total number of patients (K) are 3,532 (the HF case patients of D1 dataset). %The PARAFAC2 tensor includes a total of 290000 non-zero elements.

\subsubsection{\textbf{Hyper-parameter search:}}  
\begin{figure}[htb]
\centering
\includegraphics[scale=0.8]{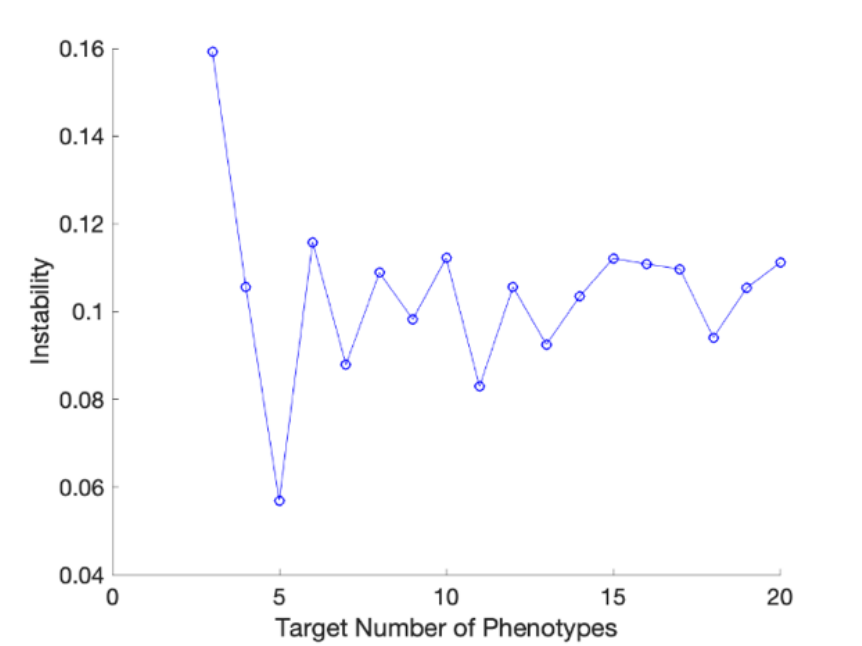}
\caption{ \footnotesize
Stability-driven metric for different values of phenotypes($R=\{3,4,...,20\}$). The optimal value of phenotypes corresponds to the minimum value of stability-driven metric ($R=5$).
}
\label{fig:opt_phen}
\end{figure}

The optimal number of phenotypes (R), $\lambda$, and $\mu$ need to be tuned. So, We perform a 3-D grid search for $R=\{3,4,5,...,20\}$, $\lambda \in \{0.01, 0.1,  1\}$ and $\mu_1 = \mu_2 = \cdots = \mu_K = \mu = \in \{0.01, 0.1, 1\}$ based on the stability-driven metric provided in \cite{wu2016stability}. Stability-driven metric calculates the dissimilarity between a pair of factor matrices from two different initialization (\M{V},\M{V}') using the cross-correlation matrix (C) as follows: $ diss(\M{V},\M{V}')=\frac{1}{2R} \Big(2R- \sum_{i=1}^{R} \underset{1 \leq j \leq R}{\max} C_{ij} -\sum_{j=1}^{R}\underset{1 \leq i \leq R}{\max} C_{ij} \Big)$ 
where diss(\M{V},\M{V}') denotes the dissimilarity between factor matrices \M{V},\M{V}'. Also cross-correlation matrix ($\M{C} \in \mathbb{R}^{R \times R} $) computes the cosine similarity between all columns of \M{V},\M{V}'.  We run \mname model for 10 different random initialization and compute the average dissimilarity for each value of $R$, $\lambda$, and $\mu$. Figure \ref{fig:opt_phen} shows the best value of stability-driven metric for different values for R.
The best solution corresponding to $R=5,  \lambda=0.1, \mu=0.1$ were chosen as the one achieving the lowest average  dissimilarity. Then, for fixed $R=5, \lambda=0.1, \mu=0.1$ we present the phenotypes of the model achieving the lowest RMSE among the 10 different runs. 

\subsubsection{\textbf{Phenotype definition:}}
We provide the details of all 5 phenotypes discovered by \mname in Table \ref{tab:other_phenotypes}.

\subsubsection{\textbf{Pure PARAFAC2 cannot handle static feature integration.}}
%\kp{i suggest we change the name of this section to: \textbf{Simpler tensor methods cannot handle static feature integration.}}

%\kp{We need to be more precise on why we have this section and what exactly we do, e.g.,: In this Section, we illustrate the fact that tweaking simpler tensor-based approaches (E.g., CP or PARAFAC2) so that they can handle static features would result in poor performance. In specific: here, let's precisely define in separate paragraphs what do we do to make CP and PARAFAC2 handle static features. The overall description here is weak: what is the dataset precisely? Also, should this section be moved after the case study? or, if the dataset is the same as the one used for our case study, maybe include it as a subsection of our case study? it is kind of weird that we mention some benefit of the phenotypes discovered by our method, but we have not yet introduced our results}
In this Section, we illustrate the fact that a naive way of incorporating static feature information into a simpler  PARAFAC2-based framework~\cite{afshar2018copa} would result in poor, less interpretable phenotypes. We incorporate the static features into PARAFAC2 input by repeating the value of static features on all clinical visits of the patients same as what we did for COPA(+static). For instance, if the male feature of patient k has value 1, we repeat the value 1 for all the clinical visits of that patient. Then we compare the phenotype definitions discovered by \mname (matrices \M{V}, \M{F}) and  by COPA (matrix \M{V}). Table \ref{tab:pheno_static_incorpor} contains two sample phenotypes discovered by this baseline, using the same truncation threshold that we use throughout this work (we only consider features with values greater than $0.1$). We observe that the static features introduce a significant amount of bias into the resulting phenotypes: the phenotype definitions are essentially dominated by static features, while the values of weights corresponding to dynamic features are closer to $0$. % This bias is reflected through the fact that due to  (i.e. in each phenotype, the static feature has a high value while the dynamic features are around zero).  
This suggests that pure PARAFAC2-based models such as the work in~\cite{afshar2018copa} are unable to produce meaningful phenotypes that handle both static and dynamic features. Such a conclusion extends to other PARAFAC2-based work which does not explicitly model side information~\cite{kiers1999parafac2,Perros2017-dh,cohen2018nonnegative}.

\begin{table}
  \centering
  \caption{Two sample phenotypes discovered by COPA(+static) baseline by naively integrating static features into a simpler PARAFAC2-based model~\cite{afshar2018copa} .}
  \scalebox{0.75}{
    \begin{tabular}{lr}
    Phenotype 1 & \multicolumn{1}{l}{weight} \\
    \midrule
    \textcolor[rgb]{ .867,  .031,  .024}{Static\_Alcohol\_yes} & \textcolor[rgb]{ .867,  .031,  .024}{0.3860} \\
    \textcolor[rgb]{ .867,  .031,  .024}{Static\_White} & \textcolor[rgb]{ .867,  .031,  .024}{0.2160} \\
    \textcolor[rgb]{ .867,  .031,  .024}{Static\_Non\_Hispanic} & \textcolor[rgb]{ .867,  .031,  .024}{0.2064} \\
    \textcolor[rgb]{ .867,  .031,  .024}{Static\_Smk\_Quit} & \textcolor[rgb]{ .867,  .031,  .024}{0.1743} \\
    \textcolor[rgb]{ .867,  .031,  .024}{Static\_male} & \textcolor[rgb]{ .867,  .031,  .024}{0.1508} \\
    \textcolor[rgb]{ .867,  .031,  .024}{Static\_moderately\_obese} & \textcolor[rgb]{ .867,  .031,  .024}{0.1025} \\
   % Dx\_Cardiac dysrhythmias & 0.10639 \\
%    Dx\_Congestive heart failure; nonhypertensive & 0.083188 \\
 %   Rx\_Coumarin Anticoagulants & 0.081391 \\
          &  \\
    Phenotype 2 & \multicolumn{1}{l}{weight} \\
    \midrule
    \textcolor[rgb]{ .867,  .031,  .024}{Static\_age\_between\_70\_79} & \textcolor[rgb]{ .867,  .031,  .024}{1} \\
    \textcolor[rgb]{ .867,  .031,  .024}{Static\_Non\_Hispanic} & \textcolor[rgb]{ .867,  .031,  .024}{0.8233} \\
    \textcolor[rgb]{ .867,  .031,  .024}{Static\_White} & \textcolor[rgb]{ .867,  .031,  .024}{0.7502} \\
    \textcolor[rgb]{ .867,  .031,  .024}{Static\_Alcohol\_No} & \textcolor[rgb]{ .867,  .031,  .024}{0.6905} \\
    \textcolor[rgb]{ .867,  .031,  .024}{Static\_moderately\_obese} & \textcolor[rgb]{ .867,  .031,  .024}{0.2098} \\
    \textcolor[rgb]{ .867,  .031,  .024}{Static\_male} & \textcolor[rgb]{ .867,  .031,  .024}{0.2026} \\
    \textcolor[rgb]{ .867,  .031,  .024}{Static\_Smk\_No} & \textcolor[rgb]{ .867,  .031,  .024}{0.1614} \\
   % Rx\_Opioid Combinations & 0.081993 \\
%    Dx\_Spondylosis; intervertebral disc disorders; other back problems & 0.05984 \\
 %   Dx\_Diabetes mellitus with complications & 0.0591054 \\
  %  Dx\_Essential hypertension & 0.058843 \\
    \end{tabular}%
    }
  \label{tab:pheno_static_incorpor}%
\end{table}%

\subsection{Recovery of true factor matrices} 
In this section, we  assess to what extent the original factor matrices can be recovered through synthetic data experiments \footnote{The reason that we are working with synthetic data here is that we do not know the original factor matrices in real data sets.}. We demonstrate that: a) \mname recovers the true latent factors more accurately than baselines for noisy data; and b) the baseline (COPA+) which does not preserve a high CPI measure (which is known to be theoretically linked to uniqueness~\cite{kiers1999parafac2}~\footnote{A discussion on uniqueness can be found in Background and \& Related Work Section.}) fails to match \mname in terms of latent factor recovery, despite achieving similar RMSE.

\subsubsection{Evaluation Metric:}  
\textbf{ Similarity between two factor matrices:} We define the cosine similarity between two vectors \V{x_i}, \V{y_j}  as $C_{ij}$ =$\frac{\V{x_i}^T\V{y_j}}{||x_i|| ||y_j||}$. Then the similarity between two factor matrices $\M{X} \in \mathbb{R}^{I \times R}$, $\M{Y} \in \mathbb{R}^{I \times R}$ can be computed as (similar to~\cite{kiers1999parafac2}):
\[
\text{Sim}(\M{X},\M{Y})=\frac{\sum\limits_{i=1}^{R} \underset{1 \leq j \leq R}{\max} C_{ij}}{R}
\]
The range of $Sim$ is between [0,1] and the values near 1 indicate higher similarity. %We use that in synthetic data experiments in Section~\ref{recovery_true_fac} to assess to what extent the ground-truth latent factors are captured by the approaches under comparison. %\kp{remember to change this one if we include average recovery instead of similarity}
%\ho{Same comment as CPI, if you do RMSE as a new line, then do all others as new lines.}

\subsubsection{\textbf{Synthetic Data Construction:}} We construct the ground-truth factor matrices $\M{\tilde{H}} \in \mathbb{R}^{R \times R}$, $\M{\tilde{V}} \in \mathbb{R}^{J \times R}$, $\M{\tilde{W}} \in \mathbb{R}^{K \times R}$, $\M{\tilde{F}} \in \mathbb{R}^{P \times R}$ by drawing a number uniformly at random between (0,1) to each element of each matrix. For each factor matrix \M{\tilde{Q}_k}, we create a binary non-negative matrix such that $\M{\tilde{Q}_k}^T\M{\tilde{Q}_k}=I$ and then compute \M{\tilde{U}_k}=\M{\tilde{Q}_k}\M{\tilde{H}}.  After constructing all factor matrices, we compute the input based on $\M{X_k= \M{\tilde{U}_k}diag(\tilde{W}(k,:))\M{\tilde{V}}^T}$  and $\M{A}=\M{\tilde{W}}\M{\tilde{F}}^T$. We set $K=100, J=30, P=20, I_k=100$, and $R=4$. We then add Gaussian normal noise to varying percentages of randomly-drawn elements ($\{5\%, 10\%, 15\%, 20\%, 25\%, 30\%, 35\%, 40\%\}$) of $\M{X}_k, \forall k=1, \dots, K$ and $\M{A}$ input matrices.

\subsubsection{\textbf{Results}:} 
\begin{figure}[t]
\centering
    \begin{subfigure}[t]{0.45\textwidth}
        \includegraphics[height=0.25 in]{results/label.pdf}
         \label{fig:label}
    \end{subfigure}\\

    \begin{subfigure}[t]{0.34\textwidth}
        \includegraphics[height=1.3 in]{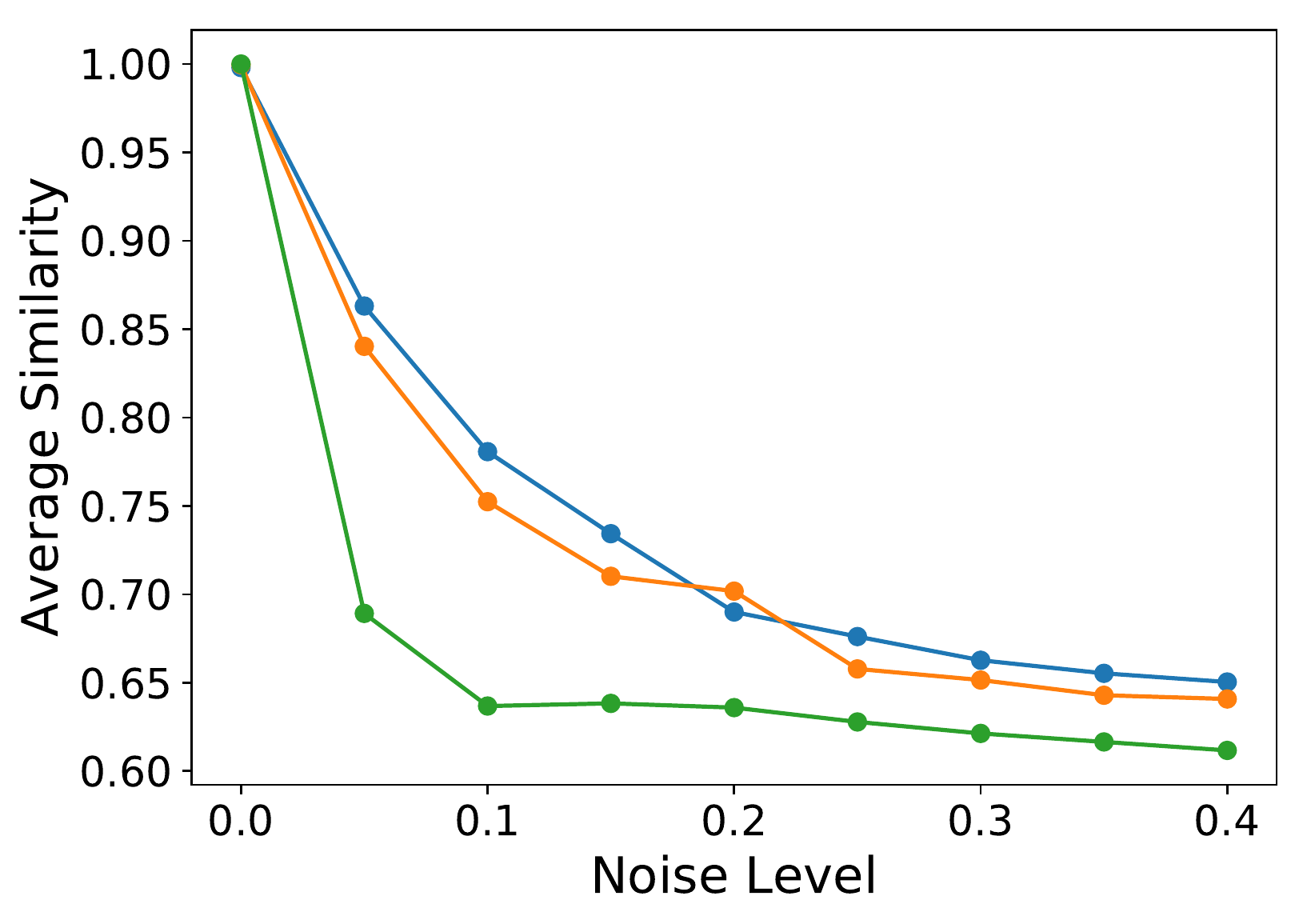}
        \caption{ Average Similarity vs Noise Level}
        \label{fig:AvgSim_CPI}
    \end{subfigure}%
    \begin{subfigure}[t]{0.34\textwidth}
        \includegraphics[height=1.3 in]{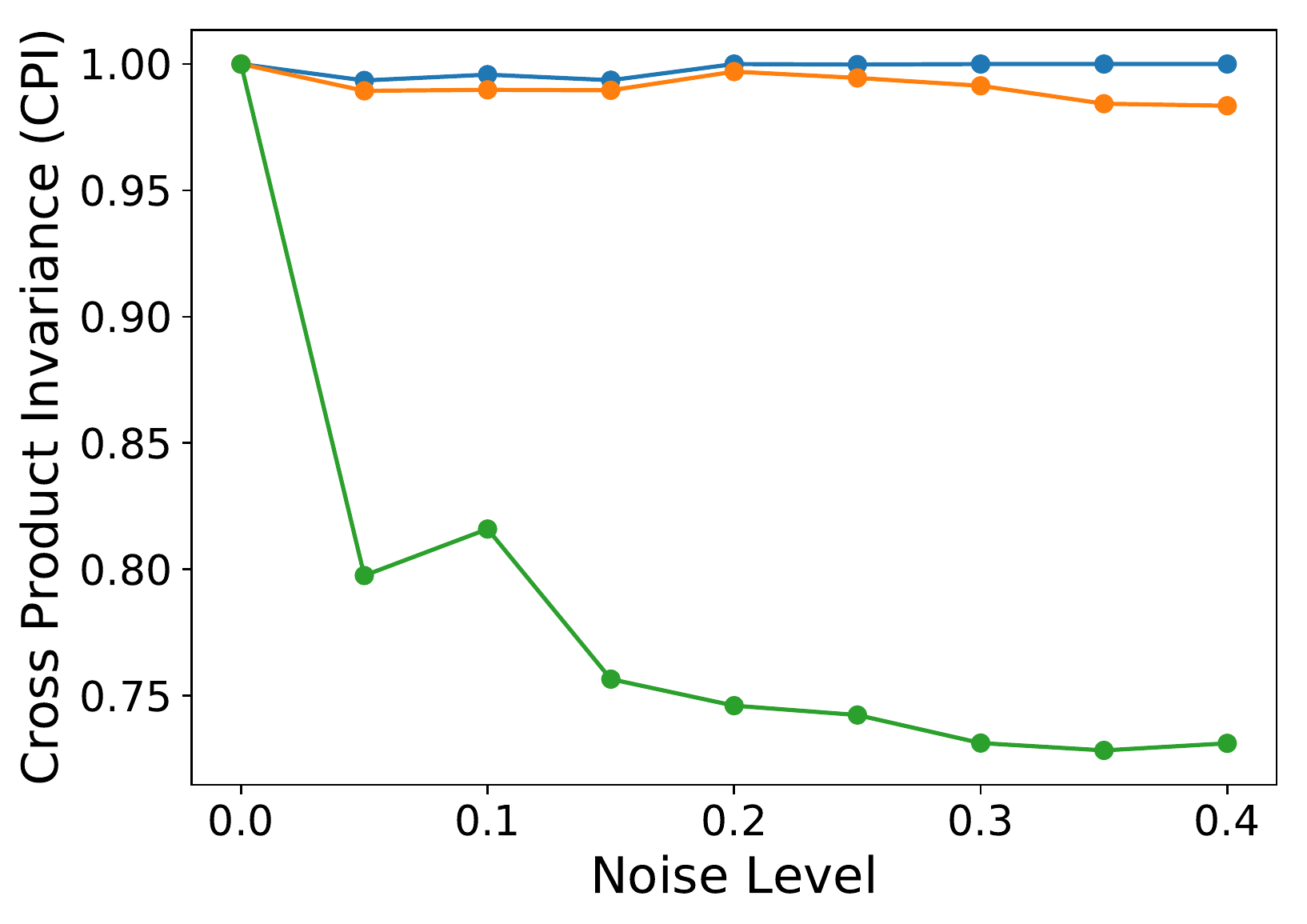}
        \caption{CPI vs Noise Level}
         \label{fig:RMSE_noise}
    \end{subfigure}\\

    \caption{  Figure~\ref{fig:AvgSim_CPI} provides  total average similarity between the estimated and the true factor matrices for different noise levels ($\{5\%, 10\%, 15\%, 20\%, 25\%, 30\%, 35\%, 40\%\}$) on synthetic data. Figure \ref{fig:RMSE_noise} provides  the CPI  of three methods  for different levels of noise  for a synthetic data with K=100, J=30, P=20, $I_k=100$, R=4. All points in the figures is computed as an average of 5 random initialization. All three algorithms achieve similar values for RMSE.}
    \label{fig:noise_metrics}
\end{figure}

For all approaches under comparison, all three methods achieve same value for RMSE, therefore, we skip ploting RMSE vs different noise levels. We assess the similarity measure between each ground truth latent factor and its corresponding estimated one (e.g., $Sim(\M{\tilde{V}}, \M{V})$), and consider the average $Sim(\cdot, \cdot)$ measure across all output factors as shown in Figure~\ref{fig:AvgSim_CPI}. We also measure CPI and provide the results in \ref{fig:RMSE_noise} for different levels of noise .  We observe that despite achieving comparable RMSE, COPA+ scores the lowest on the similarity between the true and the estimated factors. On the other hand, \mname achieves the highest amount of recovery, in accordance to the fact that it achieves the highest CPI among all approaches. Overall, 
we demonstrate how promoting uniqueness (by enforcing the CPI measure to be preserved~\cite{kiers1999parafac2}) leads to more accurate parameter recovery, as suggested by prior work~\cite{kiers1999parafac2,williams2018unsupervised}.

